\pdfoutput=1

\documentclass[11pt]{article}

\usepackage{emnlp2021}

\usepackage{times}
\usepackage{latexsym}

\usepackage[T1]{fontenc}

\usepackage[utf8]{inputenc}

\usepackage{microtype}

\usepackage{amsfonts}
\usepackage{bm}
\usepackage{array,amsmath}
\usepackage{amssymb}
\usepackage{dsfont}
\usepackage{float}
\usepackage{graphicx}
\usepackage{algorithm}
\usepackage{algorithmicx}
\usepackage{algpseudocode}
\usepackage{pgf,tikz}
\usepackage{subcaption}
\usepackage{xcolor}
\usepackage{color, colortbl}
\usepackage{bm}
\usepackage{mathrsfs}
\usepackage{multirow}
\usepackage{booktabs}
\usetikzlibrary{arrows}
\usepackage{threeparttable}
\usepackage[inline]{enumitem}
\usepackage{endnotes}

\def\figref#1{Figure~\ref{fig:#1}}
\def\figlabel#1{\label{fig:#1}\label{p:#1}}

\def\tabref#1{Table~\ref{tab:#1}}
\def\tablabel#1{\label{tab:#1}\label{p:#1}}

\def\secref#1{\S\ref{sec:#1}}
\def\seclabel#1{\label{sec:#1}}
\def\eqref#1{Eq.~\ref{eqn:#1}}

\newcounter{notecounter}
\newcommand{\enotesoff}{\long\gdef\enote##1##2{}}

\enotesoff

\long\def\eat#1{}

\title{Wine is Not v i n. \\ On the Compatibility of Tokenizations Across Languages}

\author{Antonis Maronikolakis\Thanks{ Equal contribution.} \and 
	Philipp Dufter$^*$  \and
	Hinrich Sch\"{u}tze \\
	Center for Information and Language Processing (CIS), LMU Munich, Germany\\
		{\tt {antmarakis,philipp}@cis.lmu.de}}

\begin{document}
	\maketitle
	\begin{abstract}
		
		\eat{
			The size of the vocabulary is a central design choice in
			pretrained language models with
			impact on
			both performance and memory
			requirements. Typically, subword tokenization algorithms such
			WordPiece tokenization are used. A systematic
			tuning of the vocabulary size is usually missing, and 
			there is little research on balancing a shared vocabulary across 
			many languages. However,
			learning multilingual models with stark variations in the
			tokenization is challenging. For example it would be hard
			learn a good multilingual representation of the token ``wine'' vs.\ a
			character tokenized version of the French ``v i n''. In this
			work we investigate the compatibility of tokenizations
			across languages for multilingual static and contextualized
			embedding spaces.}

		The size of the vocabulary is a central design choice in large pretrained language models, with respect to both performance and memory requirements. Typically, subword tokenization algorithms such as byte pair encoding and WordPiece are used.  In this work, we investigate the compatibility of tokenizations for multilingual static and contextualized embedding spaces and propose a measure that reflects the compatibility of tokenizations across languages.  Our goal is to prevent incompatible tokenizations, e.g., "wine" (word-level) in English vs.\ "v i n" (character-level) in French, which make it hard to learn good multilingual semantic representations.  We show that our compatibility measure allows the system designer to create vocabularies across languages that are compatible -- a desideratum that so far has been neglected in multilingual models.
	\end{abstract}

	\section{Introduction}
	
	Pretrained language
	models \cite{howard-ruder-2018-universal,peters-etal-2018-deep,devlin-etal-2019-bert}
	have become the de-facto standard in natural language processing (NLP). Due to memory constraints and to overcome sparsity, subword tokenization models such as byte pair encoding \cite{sennrich-etal-2016-neural} or WordPiece tokenization \cite{schuster2012japanese} are most commonly used. Multilingual pretrained language models (PLMs) such as mBERT \cite{devlin-etal-2019-bert} are mostly trained in an unsupervised fashion without any cross-lingual supervision. They rely on the fact that the underlying distributions across languages can be well aligned \cite{conneau-etal-2020-emerging}.

		\begin{figure}[t]
		\centering
		\def\mywidth{0.22\textwidth}
		\includegraphics[width=\mywidth]{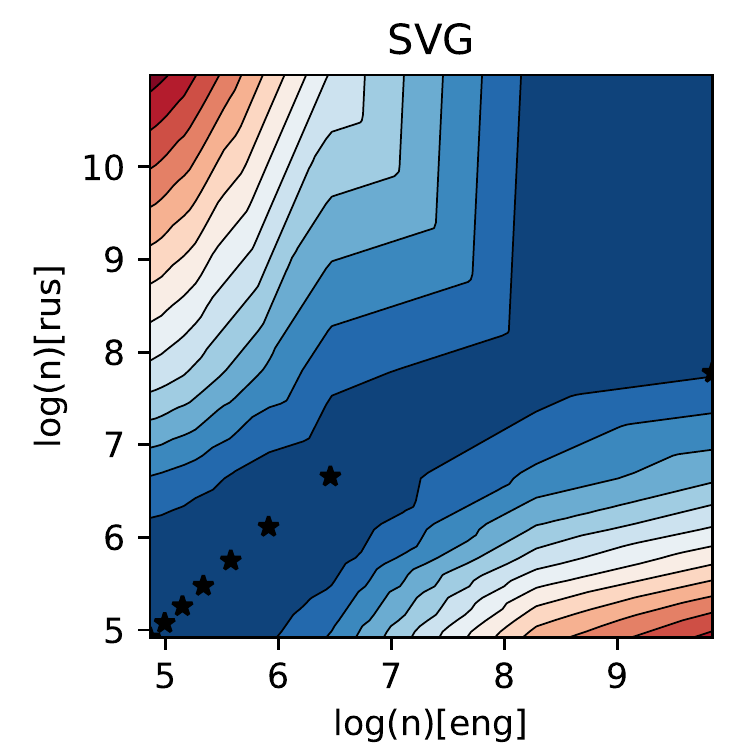}
		\includegraphics[width=\mywidth]{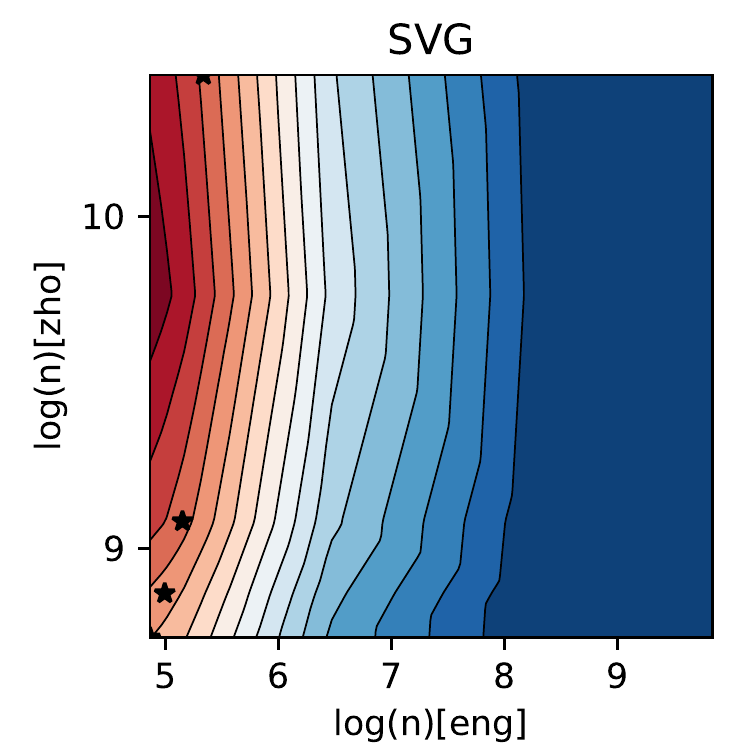}
		\caption{Compatibility of static embedding
		vocabulary sizes measured by SVG
		(singular value gap) as a function of 
                vocabulary size of English (x-axis)
		and Russian/Chinese (y-axis). 
Blue (red) indicates larger (smaller) compatibility.
Black stars indicate vocabulary sizes where absolute
		compression rates of both languages are
		equal.
                The figure confirms our intuitions about vocabulary size
		compatibility:
                for English/Russian,
similar vocabulary sizes are compatible, but for
		Chinese, we only find compatibility for
		larger English vocabulary sizes. This is
		because a Chinese logogram by itself is
		usually a meaning bearing unit, so that
		most subword (or sub-root) elements in English
		have no  analog in Chinese.
                \figlabel{pitch}}
	\end{figure}
	
	\eat{While the subword tokenization algorithms are heavily researched \cite{bostrom-durrett-2020-byte} there is little work }
	However, there is little work 
	that investigates the compatibility of tokenizations across languages. We argue that this is an important research question for several reasons.
	\begin{enumerate*}[label={\textbf{\roman{*})}}]
		\item Recent work suggests that the tokenization in multilingual models influences the multilinguality of these models \cite{rust2020how}. This corresponds to a strong intuition: in case of heavily diverging tokenizations, e.g., word tokenization in English and character tokenization in French it will be hard to train common cross-lingual representations.
		\item Given a limited vocabulary budget, e.g., from memory constraints, a researcher is naturally interested in how to distribute the budget across multiple languages.
		\item Currently, a balance of the
	tokenization is heuristically achieved through up-
	and downsampling
	(e.g., \citealp{conneau-etal-2020-unsupervised}). With
	a better tokenization strategy, one might be able to achieve performance improvements.
	\end{enumerate*}
	
	Research in multilinguality has so far
	focused, among other directions, on analyzing and
	improving cross-lingual
	transfer \cite{artetxe-etal-2020-cross,K2020Cross-Lingual,wu-dredze-2019-beto,conneau-etal-2020-emerging}
	and modeling
	tasks \cite{conneau_cross_lingual,huang-etal-2019-unicoder}. Earlier
	work on tokenization focused on the development of
	algorithms \cite{schuster2012japanese,gage_bpe}
	while analysis of the effect of tokenization has
	been undertaken both in the general \cite{bostrom-durrett-2020-byte} and in the multilingual domain \cite{wei2021training}.
	
	We propose to investigate the compatibility of tokenizations across languages systematically. To this end, we train both static and contextualized embeddings using different vocabulary sizes and WordPiece tokenization \cite{schuster2012japanese}. Subsequently we investigate the similarity of the embedding spaces and their degree of multilinguality. 
	
	\figref{pitch} shows an example of our research
	question: here we trained static embedding spaces
	and investigated the similarity of the embedding
	spaces by computing the singular value gap (SVG) measure. The plots show that different vocabulary sizes yield very different similarities. In addition, language pairs behave quite differently. 
See the caption for more information.

	\eat{Specifically, we examine bilingual models, pairing English with Greek, Russian and Chinese. These languages were chosen because they exhibit different morphological and grammatical complexities. This gives us ample room for investigating the compatibility of tokenizations across linguistically diverse languages.}
	
	In summary, our contributions are:\footnote{Code available at \url{https://github.com/antmarakis/vocab_size_compat}}
	\begin{enumerate}
		\item We propose an experimental setting that allows the comparison of embeddings at different tokenization levels with each other.
		\item We provide evidence that vocabulary size compatibility matters, i.e., if tokenizations across languages are incompatible it is hard to achieve multilingual representations.
		\item We propose a measure that indicates the compatibility of tokenizations.
	\end{enumerate}

	\section{Related Work}
	
In a position paper, \citet{artetxe-etal-2020-call}
advocated for more rigor in cross-lingual research. The
topic of tokenization was raised, especially the fact that
current word and subword tokenization algorithms do not
adequately capture morphological nuances and do not do
justice to the great variety of languages in the world like those that use logographic scripts. These considerations motivate us to propose a method that makes better use of vocabulary space, a facet of  multilingual models that is often overlooked.

\subsection{Multilingual Research}

Investigation of the multilinguality of BERT has gathered momentum in recent years. \citet{wu-dredze-2019-beto,nooralahzadeh-etal-2020-zero,artetxe-etal-2020-cross,conneau-etal-2020-emerging} investigate cross-lingual transfer, with an extensive study of multilinguality at-scale presented in \citet{conneau-etal-2020-unsupervised}. \citet{K2020Cross-Lingual} research how linguistic variation in language properties affects multilingual BERT models, with \citet{dufter-schutze-2020-identifying} further investigating the effect of model architecture and linguistic properties on BERT's multilinguality. \citet{conneau_cross_lingual} focus on variations of modeling tasks and their effect on downstream tasks such as XNLI \cite{conneau2018xnli} and machine translation. A novel set of cross-lingual pre-training tasks was introduced in \citet{huang-etal-2019-unicoder}. \citet{anastasopoulos-neubig-2020-cross} show that performance in bilingual experiments is not optimal when English is used as the ``hub'', but it depends on the language pair.

Work on the representation space of BERT has also been abundant. \citet{singh-etal-2019-bert} investigate the representation space of mBERT and its properties and \citet{pires-etal-2019-multilingual} initially researched the degree of multilinguality of mBERT.

In our work, we further investigate the zero-shot
        capabilities of bilingual models and
        present  a method for comparing multilingual embeddings across tokenizations.

\subsection{Tokenization Research}

Another vein of research into model analysis has
been the study of the effect of
tokenization. \citet{heinzerling-strube-2018-bpemb}
compute BPE embeddings for 275 languages and perform
a cross-lingual study on various tokenization
methods. Further study into BPEs is conducted
by \citet{wei2021training}, where byte-level BPEs
are compared against their character-level
counterparts as it pertains to multilingual models
while BPE vocabulary size is examined in the context
of neural machine translation
by \citet{gowda-may-2020-finding}. 
\citet{bostrom-durrett-2020-byte} show that
byte-pair encoding is suboptimal when pretraining language
models.  In our work, we investigate
another prevalent tokenization method, WordPiece, and show
that vocabulary size plays an important role in achieving
multilinguality in BERT.
	
Research into the selection of tokenization algorithms has
produced alternative methods. For example,
\citet{kudo-2018-subword} present a subword
regularization method where a model is trained with multiple
tokenizations in order to reinforce robustness in the model;
and \citet{aguilar2020char2subword} bridge the
gap between subword and character-level models, proposing a
module that learns to approximate subword embeddings given
characters. \citet{asgari2021subword,provilkov-etal-2020-bpe} further research subword tokenization methods. With our work we introduce a measure that can
help NLP researchers pick vocabulary sizes with a more
robust procedure.

\section{Vocabulary Sizes}
\seclabel{vocabsizes}

\subsection{Tokenization}

Let $\mathcal{C}$ be the set of unicode code points. 
We define a tokenizer as a function $\tau: \mathcal{C}^{t} \to \mathcal{V}_{\tau}^s$ that maps a sequence of $t$ unicode code points to a sequence of $s$ vocabulary tokens. $\mathcal{V}_{\tau}$ is the vocabulary of the tokenizer and $n_{\tau} := |\mathcal{V}_{\tau}|$ is the vocabulary size of the tokenizer.

Usually the tokenizer is trained on a corpus $\mathcal{U}$. 
Popular tokenizers are byte-pair encoding \cite{sennrich-etal-2016-neural}, WordPiece \cite{schuster2012japanese} and SentencePiece \cite{kudo-richardson-2018-sentencepiece}. An example tokenization using the WordPiece tokenizer of the PLM bert-base-multilingual-cased for the input $c = \text{``Exceptional weather.''}$ is $\tau(c) = [\text{``Ex''}, \text{``\#\#ception''}, \text{``\#\#al''}, \text{``weather''}, \text{``.''}]$, where ``\#\#'' is a continuation symbol.

Throughout this paper we use the WordPiece tokenizer. The main motivation is that important successful PLMs use this tokenization method. In future work we plan to extend this analysis to other tokenization algorithms as well.

\subsection{Compression Rate}

\begin{figure}[t]
\centering
\def\mywidth{0.475\textwidth}
\includegraphics[width=\mywidth]{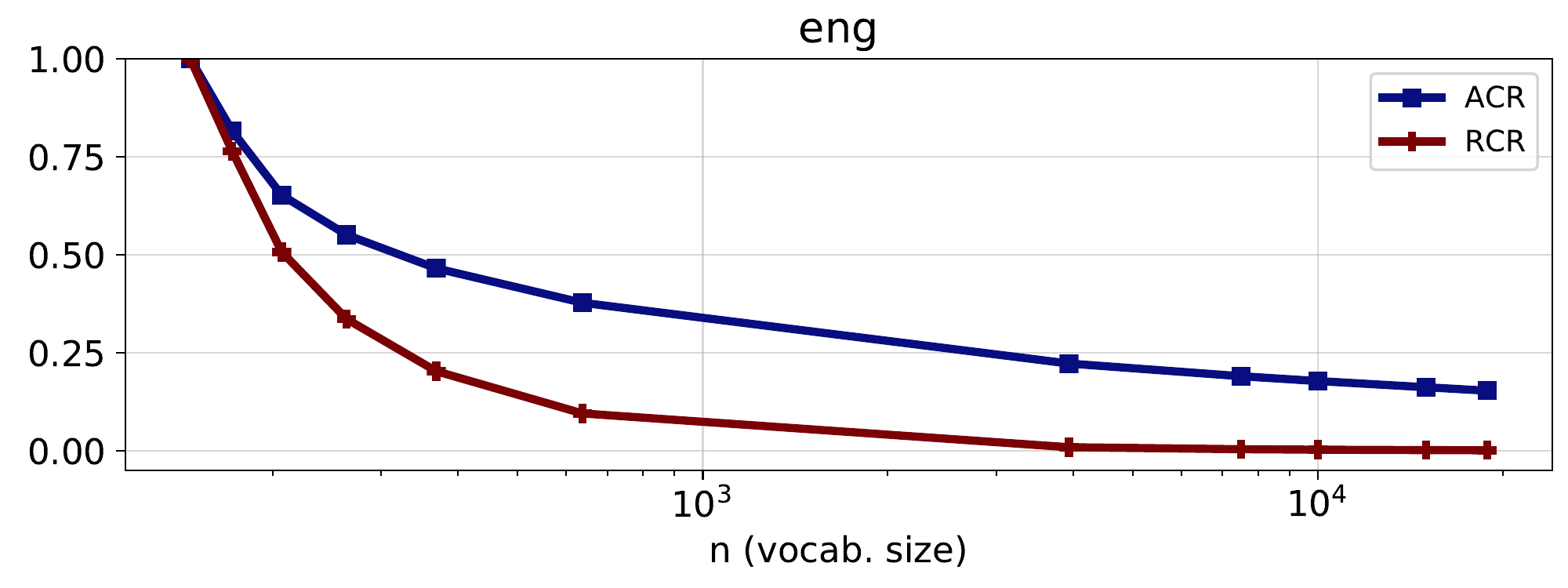}
\includegraphics[width=\mywidth]{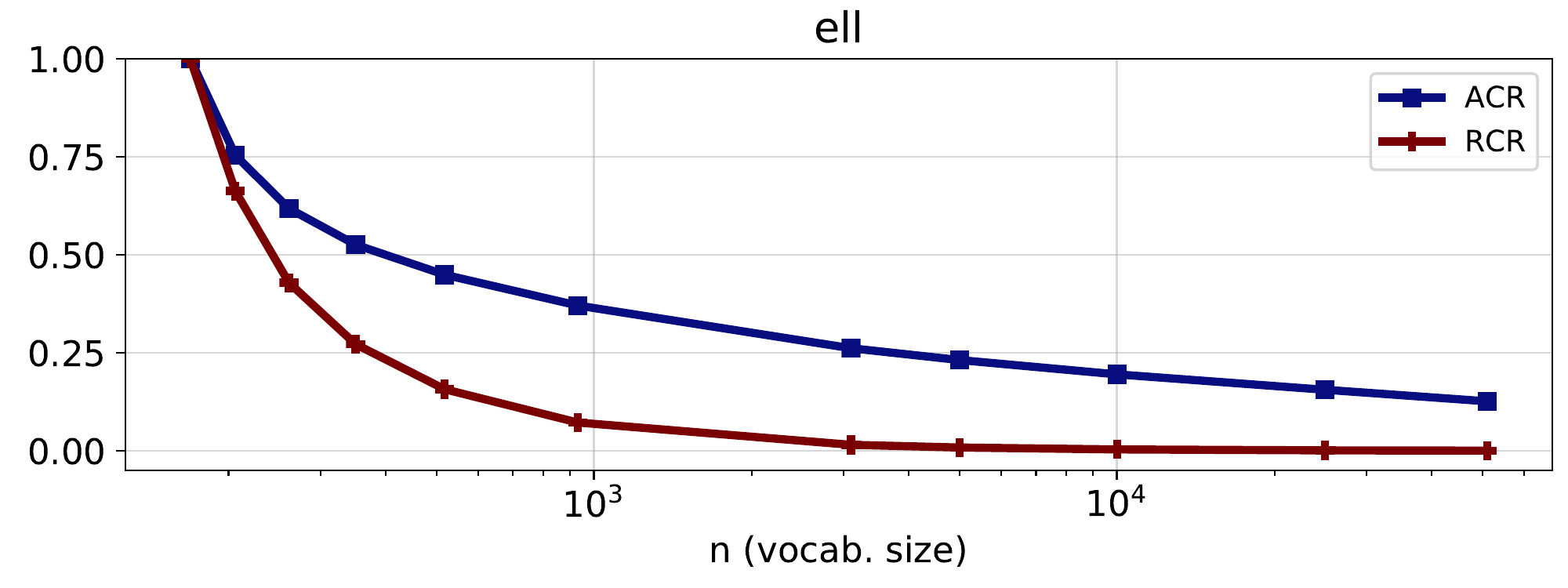}
\includegraphics[width=\mywidth]{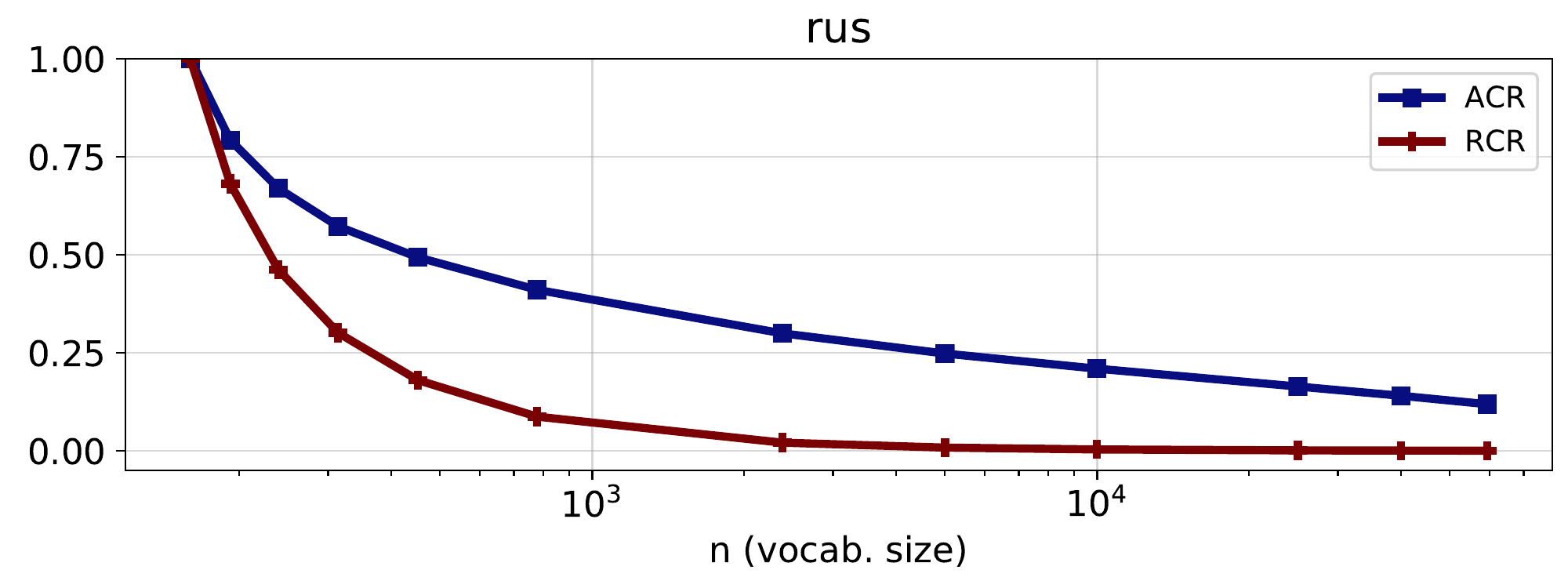}
\includegraphics[width=\mywidth]{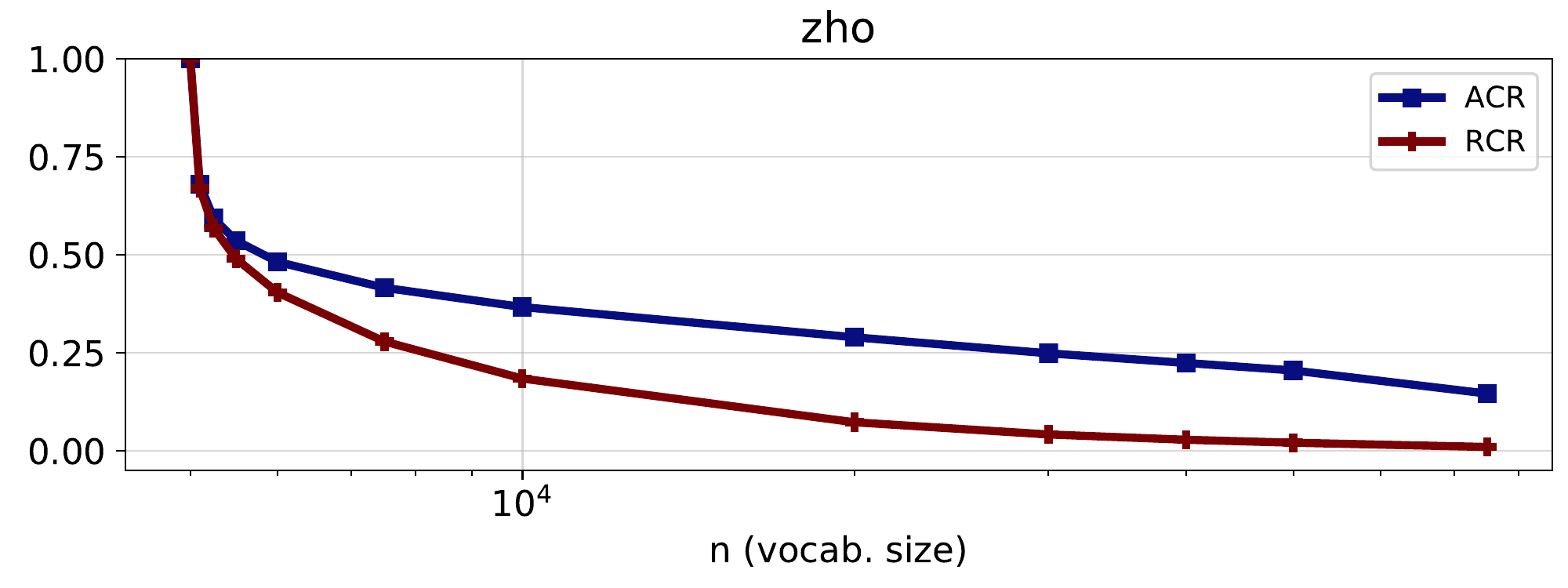}
\caption{Compression rates for four languages and different vocabulary sizes. One can see that Chinese using logograms has a much steeper curve.\figlabel{compression}}
\end{figure}

\citet{rust2020how} use two measures to compare tokenizations across languages. However, they assume that each language has the notion of ``words'' that are separated by whitespace. For example they compute the proportion of words that are split into multiple tokens by the tokenizer. 

We posit the following
desiderata
for a measure that compares tokenizations across languages:
\begin{enumerate*}[label={\textbf{\roman{*})}}]
\item Assume that a text consists of a sequence of unicode code points. Do not require the existence of ``words''.
\item Take into account vocabulary size and alphabet size. 
\item Take into account the frequency of tokens in a corpus.
\end{enumerate*}

The first two
desiderata
should ensure that the measure is
as inclusive as possible with regard to the variety of
languages; the last
desideratum
ensures that the
measure considers how ``important'' different tokens are in
a corpus. Based on these desiderata, we  propose two measures, an absolute and
a relative compression rate.

\textbf{ACR $r^{ABS}$.}
Let $c \in \mathcal{C}^{t}$ be a corpus with $t$ unicode code points and $\tau_n(c)$ be the tokenized version of the corpus for a tokenizer with vocabulary size $n$. The absolute compression rate is then simply
$$
r^{ABS}(n) = \frac{|\tau_n(c)|}{|\tau_{n_{\min}}(c)|},
$$
where $\tau_{n_{\min}}$ is the tokenizer with minimal vocabulary size $n_{\min}$. Note that the tokenization provided by $\tau_{n_{\min}}$ can be different from character tokenization and highly depends on the behavior of the tokenizer. For example for the WordPiece implementation that we use\footnote{We use \url{https://github.com/huggingface/tokenizers}} we obtain all unicode code points twice in the vocabulary, once with a continuation symbol and once without.

\textbf{RCR $r^{REL}$.}
One potential disadvantage of ACR is that it does not take into consideration how many different unicode code points are used. Thus we introduce the relative compression rate that considers in addition how ``inflated'' the vocabulary is. 
$$
r^{REL}(n) = \frac{|\tau_n(c)|}{|\tau_{n_{\min}}(c)|} \frac{n_{\min}}{n} .
$$
One can also interpret the measure as a relative inverse type-token ratio
$$
r^{REL}(n) = \frac{|\tau_n(c)|}{n} / \frac{|\tau_{n_{\min}}(c)|}{n_{\min}} .
$$
The measures have convenient properties. For example they are monotonically decreasing and $r^{ABS}(n_{\min}) = r^{REL}(n_{\min}) = 1.0$. In the WordPiece implementation that we use the maximum tokenization length is limited by the number of whitespaces in the text. Thus $r^{ABS}$ is always above a language specific threshold $a$ and $r^{ABS}(n_{\text{whitespace}}) = |\tau_{n_{\text{whitespace}}}(c)|  / |\tau_{n_{\min}}(c)| =: a$. For other tokenizers that treat whitespace as a normal unicode code point the whole text might be considered a single token, provided that the vocabulary size is sufficiently large, and thus $r^{ABS}(n_{\max}) = 1  / |\tau_{n_{\min}}(c)| =: a$. For the relative compression rate one can see that $r^{REL}(\infty) = 0.0$.

\figref{compression} shows both compression rates for four languages as computed on the PBC.

\subsection{Vocabulary Size Selection}\seclabel{vss}

In our study we want to compare different vocabulary sizes. However, it is not trivial to come up with meaningful vocabulary sizes. Uniformly chosen sizes are not interesting as increasing the vocabulary size from, e.g., 15,000 to 16,000 in English has only a marginal effect on the actual tokenization as it mostly affects rare words. Similarly, a vocabulary size of 1000 can be decent in English, but is most likely way too small for Chinese. 

We thus try to model the function $r^{ABS}$ for each language in order to be able to assess how important different vocabulary sizes are. To this end, we sample $k$ vocabulary sizes uniformly from $[n_{\min}, 100000]$ and obtain data pairs $[(n_1, r^{ABS}(n_1)), \dots, (n_k, r^{ABS}(n_k))]$. We observe that the compression rate follows an exponential pattern and thus we assume the following statistical model
$$
y:=r_\theta(x) = \alpha (x / n_{\min})^\beta + \gamma,
$$
where $\theta= (\alpha, \beta, \gamma)$. Based on the asymptotic behavior of our function we assume $\gamma = a$ and  $\alpha = 1 - a$. This gives us $r_\theta(x) = (1 - a) (x / n_{\min})^\beta + a$ and the only parameter is $\beta$.

We fit $\beta$ with a simple linear regression without intercept, i.e., $\log\left(\frac{(y - a)}{(1 - a)}\right) = \beta \log\left(\frac{x}{n_{\min}}\right)$. Overall we obtain an estimate $\hat{\beta}$ and thus a model  $r_{\hat{\beta}}^{ABS}$.

We can now sample interesting vocabulary sizes e.g., by sampling in areas where the rate of change 
$(r_{\hat{\beta}}^{ABS})'$ is high. Alternatively, we go uniformly through possible compression rates: starting from a maximum compression rate we select values at 0.1-mark intervals (1.0, 0.9, \dots) until we reach the minimum compression rate. Using $(r_{\hat{\theta}}^{ABS})^{-1}$ we obtain  vocabulary sizes from the chosen compression rates.

\tabref{table:vocabsizes} shows selected vocabulary sizes for each language.

\begin{table}[]
		\centering
		\footnotesize
		\begin{tabular}{|c|c|c|c|}
			\hline
			\cellcolor[HTML]{B5B5B5} ENG & \cellcolor[HTML]{B5B5B5} ELL & \cellcolor[HTML]{B5B5B5} RUS & \cellcolor[HTML]{B5B5B5} ZHO \\ \hline
			$129_{1.0}$ & $143_{1.0}$ & $137_{1.0}$ & $6189_{1.0}$ \\ \hline
			$147_{0.9}$ & $167_{0.9}$ & $161_{0.9}$ & $7054_{0.9}$ \\ \hline
			$172_{0.8}$ & $201_{0.8}$ & $195_{0.8}$ & $8733_{0.8}$ \\ \hline
			$209_{0.7}$ & $251_{0.7}$ & $244_{0.7}$ & $17083_{0.7}$ \\ \hline
			$267_{0.6}$ & $330_{0.6}$ & $323_{0.6}$ & $33786_{0.68}$ \\ \hline
			$374_{0.5}$ & $475_{0.5}$ & $471_{0.5}$ & - \\ \hline
			$653_{0.4}$ & $821_{0.4}$ & $827_{0.4}$ & - \\ \hline
			$4139_{0.3}$ & $2524_{0.3}$ & $2632_{0.3}$ & - \\ \hline
			$18865_{0.29}$ & $51077_{0.24}$ & $59341_{0.24}$ & - \\ \hline
		\end{tabular}
		\caption{Selected vocabulary sizes for each language. Subscript shows ACR.}
		\tablabel{table:vocabsizes}
	\end{table}

	\eat{
		\subsection{WordPiece Tokenization}
		
		In this paper, we use WordPiece \cite{schuster2012japanese}, a member of the family of automatic tokenization algorithms that are widely used in deep learning models. This algorithm begins with character-level tokens and iteratively combines and adds to the vocabulary the most frequent tokens until a target vocabulary size is reached. One of the main advantages of this method is that the issue of out-of-vocabulary words is mitigated, since the algorithm operates on the subword level. WordPiece has become prevalent in modern Deep Learning research, especially work surrounding transformers. Alongside the Byte-Pair Encoding algorithm \cite{sennrich-etal-2016-neural,gage_bpe}, WordPiece has become an indispensable block in the NLP toolkit.
		
		Despite its popularity, not a lot of care is being taken when choosing vocabulary sizes and it is an open question whether optimal or more efficient ways of selecting sizes exist. This lack of understanding may be a covert contributor to suboptimal performance. In the multilingual setting, where tokenizations vary greatly, making use of a potential vocabulary size budget is pivotal in training models both efficiently and effectively.}

	\section{Experiments}
	
	\subsection{Data}
	
	The Parallel Bible Corpus (PBC) by \citet{mayer-cysouw-2014-creating} is a multi-parallel corpus spanning 1259 languages and up to 30k verses per translation. We chose this corpus for three reasons:
	\begin{enumerate*}[label={\textbf{\roman{*})}}]
		\item For a clean experimental setup in our sentence retrieval task we require a multi-parallel corpus with additional paraphrases within the same language. This is possible in PBC as there are multiple translations for languages such as ENG, ELL or RUS.\footnote{Language codes follow the ISO 639-3 standard: \url{https://iso639-3.sil.org/}}
		\item The overall corpus size is small and
	allows for efficient training of multilingual static
	and contextualized
	models. \citet{dufter-schutze-2020-identifying}
	provided evidence that experimenting on a small setup and verifying key results in a larger setup is feasible.
		\item The corpus has a vast language coverage.
	\end{enumerate*}

	We compute some general statistics (see Appendix \tabref{table:pbcstats}) on PBC, calculating the maximum, average and minimum verse character length for each examined language, as well as the length covering 95\% of all instances. This information was used in conjunction with computational considerations to set the maximum sequence length for our experiments.
	
	\subsubsection{Language Selection}
	
	We perform our static embedding analysis on all 61 complete ``newworld'' editions in the PBC. These are very literal translations spanning the same set of verses across languages and thus ensure high comparability. For contextualized representations, experiments are more expensive. Thus we chose to focus on languages ENG, ELL, RUS and ZHO, which span different scripts, degrees of morphology and language families.
	
	\subsubsection{Vocabulary Size Selection}
	
	Instead of hand-picking arbitrary sizes without guidance behind the selection process we follow our proposal from \secref{vss} and select sizes based on compression rates. Starting from a maximum compression rate (i.e., smallest vocabulary size) we choose sizes at 0.1-mark intervals (1.0, 0.9, \dots) until we reach a minimum compression rate (i.e., maximum vocabulary size).

	\subsubsection{Fake-English}
	\seclabel{subsec:fakeenglish}
	
	We also experiment using Fake-English, denoting the English to Fake-English pair by $EngFake$. Fake-English is a concept examined in \citet{K2020Cross-Lingual}, where English characters are converted to some separate, special characters such that the model processes English and Fake-English as two different languages. The ``new'' data that is created exhibits the same grammar and structure as English, but the script is different. This offers both a simple upper performance limit and a scrying glass into potential bugs or shortcomings of our approach.
\eat{	 Further, since performance in $EngFake$ experiments is expected to be high \cite{K2020Cross-Lingual,dufter-schutze-2020-identifying}, if we do not reach such high accuracy consistently, we can assume there is something wrong with our setup or code. This can speed up prototyping and debugging by a great deal. We generate Fake-English by adding the special `0en0' symbol in front of each token. As an example, the English text `purple haze' would get converted to (for example) `0en0pur0en0ple 0en0ha0en0ze'. Since `0en0' does not occur naturally in the data, there will be no tokenization errors. }
	 Note that to generate the Fake-English data, we used the same edition as the English Bible set, to ensure better comparability.
	
	\subsection{Static Embeddings}
	
	\subsubsection{Training}
	
	We evaluate vocabulary compatibility in static embedding spaces. To this end we tokenize comparable corpora using learned WordPiece vocabularies with different sizes. Subsequently we obtain static embeddings using fastText \cite{bojanowski-etal-2017-enriching} with default parameters.
	
	\subsubsection{Compatibility Evaluation}
	
	Comparing static embedding spaces at different
	granularities is challenging. Tasks like word
	translation or sentence retrieval disqualify, as
	obtaining word or sentence representations through
	mean pooling is too crude to get discriminative
	evaluation results. Training a neural network on top of the embeddings to evaluate multilinguality requires additional input, might obliterate the effect we want to investigate (i.e., the compatibility of the embedding spaces) and is not feasible when processing many language pairs. Therefore we employ a popular method to predict the cross-lingual performance using the spectral similarity of embedding spaces. \citet{dubossarsky-etal-2020-secret} introduced the measure \emph{singular value gap} (SVG) and showed that it correlates well with cross-lingual downstream performance such as bilingual lexicon induction.
	Given two embedding spaces for two languages $\mathbf{X}_{l}\in \mathbb{R}^{n_l\times d}$ and $\mathbf{X}_e\in \mathbb{R}^{n_e\times d}$ it is computed as 
	$$
	SVG(\mathbf{X}_{l}, \mathbf{X}_e) := \sum_{i = 1}^{d}  \left(\log(\sigma_i^{e}) - \log(\sigma_i^{f})\right)^2
	$$
	with $(\sigma_i^{k})$ being the sorted singular
	values of $\mathbf{X}_{k}$. Instead of computing the
	sum across all dimensions we
	follow \cite{dubossarsky-etal-2020-secret} and
	consider only the first 40 singular values. We found
	that experiments with the two alternative measures
	cond-hm and econd-hm
	that \citet{dubossarsky-etal-2020-secret} suggest yield similar results and thus focus only on SVG fo the sake of clarity.

	\subsection{Contextualized Embeddings}
	\seclabel{subsec:bibleexps}
	
	\subsubsection{Pretraining}
	
	\textbf{Model Details.}
	For our experiments, we made use of
	BERT \cite{devlin-etal-2019-bert}. We performed two
	separate sets of experiments, one on PBC and another
	on the Wikipedia corpus.
        The Wikipedia experiments serve as a verification of the generalization of our findings to a larger scale.
	
	A thorough and rigorous analysis on the effect of vocabulary sizes can be prohibitively costly on the larger scale. When comparing the vocabulary sizes of two languages with $m$ and $n$ different sizes each, we need $m \times n$ different models. For example, to compare English ($m = 9$) with Greek ($n = 9$) we need to train $9 \times 9 = 81$ models. Pretraining all these models on Wikipedia with the default BERT parameters is prohibitively expensive.
	
	Thus, we split our experiment into the small-scale Bible setup and the large-scale Wikipedia setup (\secref{subsec:wikiexps}). We also downsized the models for the Bible experiments. These models use a single attention head, which has been shown to have  competitive performance compared to the multi-head approach with a substantial drop in required computational resources \cite{NEURIPS2019_2c601ad9}. We also downsized the hidden layer size, from 3072 to 256. We call this model BERT$_{bible}$.
	
	For the Wikipedia experiments, the default BERT hyperparameters were kept. We use BERT$_{wiki}$ to denote this model architecture.
	
	\textbf{Bible Experiments.}
	For our work, we pretrained the downsized BERT$_{bible}$ models on the PBC data. Then we evaluated on a bidirectional sentence retrieval task after finetuning following the SBERT example \cite{reimers-2019-sentence-bert}.
	
	We select the English (ENG), Greek (ELL), Russian (RUS) and Chinese (ZHO) versions of the Bible. Development and test sets were defined using common verses across the languages. The training set varies slightly for each language, since we included all verses not in the dev/test sets. Details on these splits can be found in \tabref{table:versesizes}.
	
	For pretraining, we pair ENG with ELL/RUS/ZHO. We denote these pairs with $EngEll$, $EngRus$ and $EngZho$. For each pair, we take the Cartesian product of the corresponding vocabulary sizes. The training sets for the two languages are then concatenated and two tokenizers are trained with the examined vocabulary sizes. Model pretraining then takes place for 75 epochs using BERT$_{bible}$. Batch size is set to 128 and the maximum sequence length is also set to 128. We used ADAM \cite{adam}, with a learning rate of 2e-3, a weight decay of 0.01 and an epsilon value of 1e-6. Three runs were performed and the results averaged.
	
	\subsubsection{SBERT Finetuning}
	
	Each pretrained model is finetuned under the SBERT paradigm \cite{reimers-2019-sentence-bert}. The model learns to identify whether two sentences are similar or not. Sentence representations are computed as usual and then pushed through a pooling layer. These representations are compared with cosine similarity, with the mean-squared-error loss used as the objective function.
	
	The main motivation for this finetuning is to obtain meaningful sentence representations from different tokenization granularities. Overall, this allows us to analyze the multilinguality of the representations even though the tokens in both languages might be very different, such as words and characters. During finetuning, the model also learns how to separate dissimilar sentences and bring similar ones closer together in the representation space, which allows us to evaluate multilinguality using cross-lingual sentence retrieval.

	The data used for this task comes from the PBC
	development sets. For each language we use two
	different editions, that is, two separate
	translations in the same language, i.e.,
	paraphrases. Across the two editions, the verses
	with the same identifier are paired and labeled as
	similar, while for the dissimilar examples we pair
	each verse from one edition to a different, random
	verse in the other edition. We keep this pairing
	within each language, to ensure that  our
	cross-lingual evaluation setup is zero-shot.

	\subsubsection{Compatibility Evaluation: Sentence Retrieval}
	
	To evaluate these pretrained and finetuned models, we use a cross-lingual sentence retrieval task. Data for this evaluation task comes from the development sets for English and the paired language. Given an English verse, we are tasked with retrieving the corresponding verse in the paired language. A mean-pooling layer takes as input the token representations of a sentence after a forward pass and outputs the final verse representation. Then, for each English sentence, we retrieve the 10 most similar sentences from the paired language, as calculated via cosine similarity. Our evaluation measure is precision@10. This method was repeated analogously in the other direction: for each sentence in the paired language, we retrieve the most similar sentences in English. Scores from both directions are averaged and results are analyzed in \secref{subsub:bibleexps}.
	
	Note that the cross-lingual sentence retrieval is zero-shot in the sense that any multilinguality can only stem from the pretraining as no cross-lingual information was used during SBERT finetuning.

	\section{Results}
	
	\subsection{Static Embedding Learning}
	\seclabel{staticisoresult}
	
	\begin{figure}[t]
		\centering
		\def\mywidth{0.23\textwidth}
		\includegraphics[width=\mywidth]{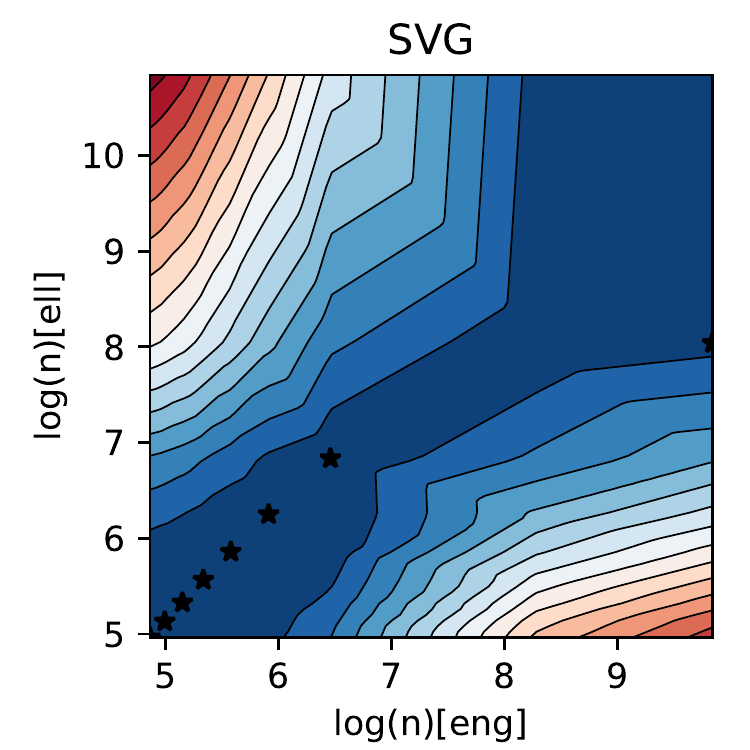}
		\includegraphics[width=\mywidth]{assets/plots3,12,svg/12,eng_newworld2013,rus_newworld}
		\includegraphics[width=\mywidth]{assets/plots3,12,svg/12,eng_newworld2013,zho_newworld}
		\includegraphics[width=\mywidth]{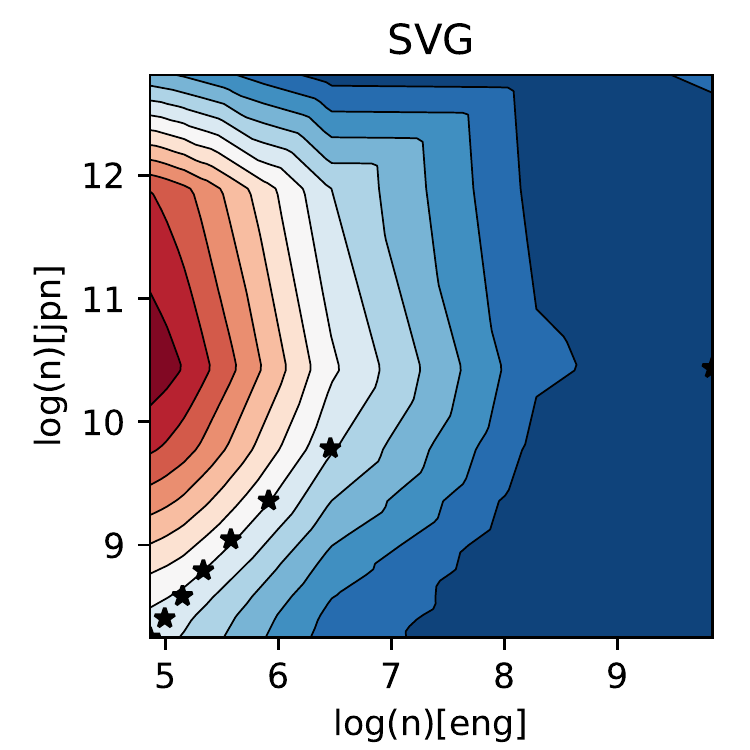}
		\includegraphics[width=\mywidth]{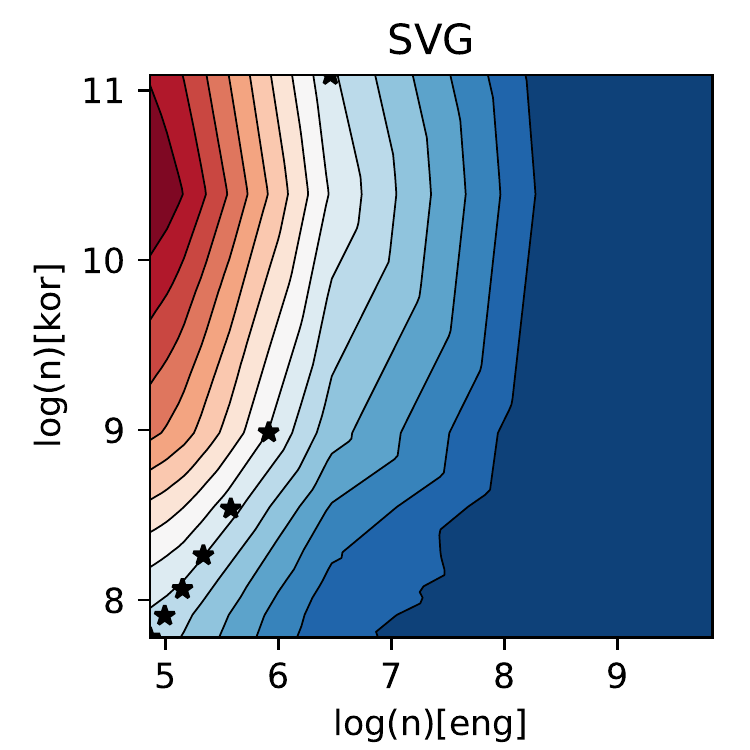}
		\includegraphics[width=\mywidth]{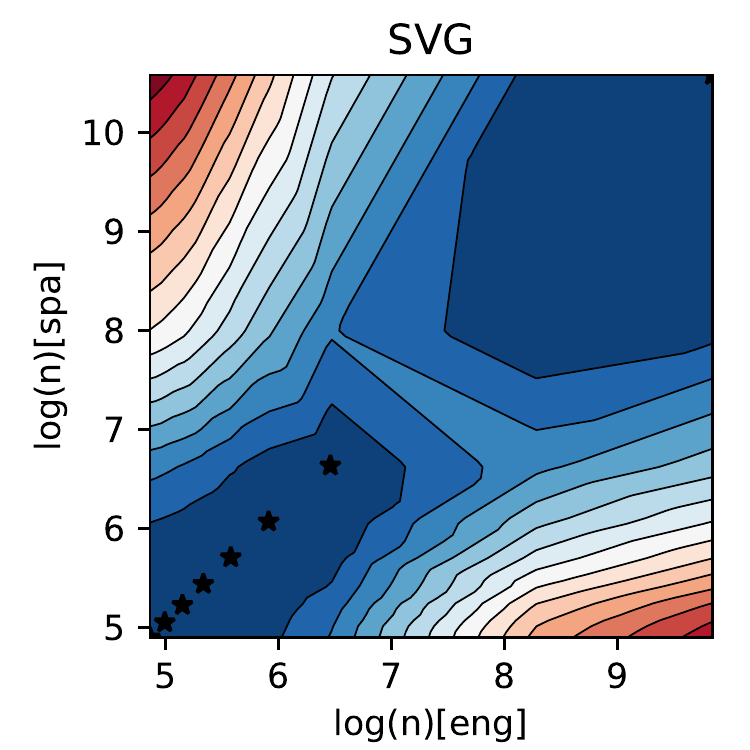}
		\caption{Figures showing the interpolated SVG of static embedding spaces for different language pairs computed on the PBC. Blue indicates high compatibility (i.e., low SVG) and red the contrary. Black stars indicate vocabulary sizes where $r^{ABS}$ of both languages are equal. One can see different patterns: alphabetic languages tend to favor the diagonal. Logograms require much larger vocabulary sizes in English. Japanese uses a mixture of logograms and a syllabic writing system and thus exhibits a special pattern. \enote{pd}{todo for myself: add results from new testament only} \figlabel{spectral}}
	\end{figure}
	
	We show results for static embeddings in \figref{spectral}. For better visualization we interpolated the results using LinearTriInterpolator in the Matplotlib library \cite{Hunter:2007}. For alphabetic languages it is favorable to have comparable vocabulary sizes, which can be seen from the clear diagonal pattern of ELL, RUS and SPA. Chinese, being based on logograms, requires large vocabulary sizes in English. This makes intuitive sense: with small vocabulary sizes in English one arrives at a character tokenization and it is hard to imagine the equivalent of a Latin character or character n-gram in Chinese. Japanese exhibits a unique pattern, which can be explained by the fact that Japanese uses both syllables and logograms. Plots for all languages can be found in Appendix \ref{app:svg_plots}.

	\begin{figure}
		\centering
		\def\mywidth{0.23\textwidth}
		\includegraphics[width=\mywidth]{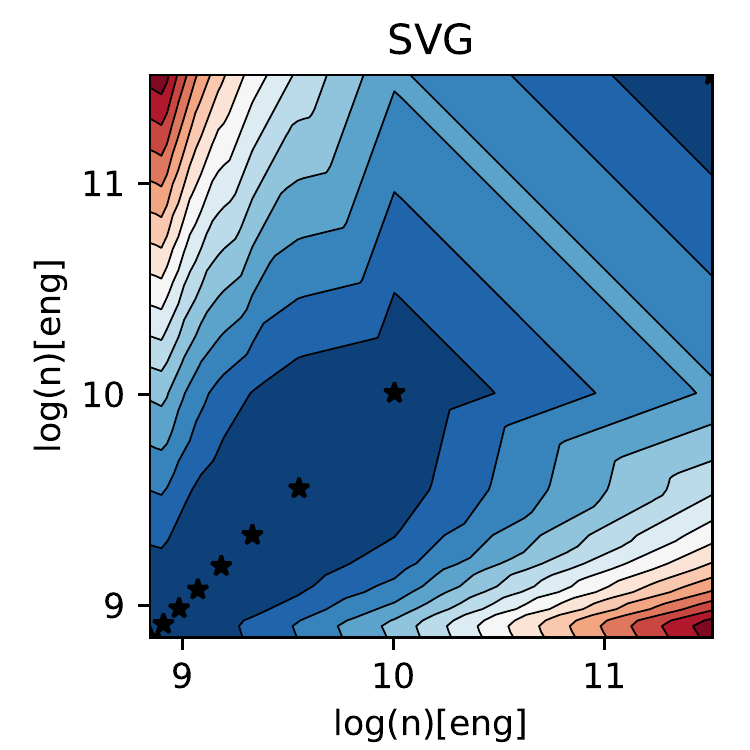}
		\includegraphics[width=\mywidth]{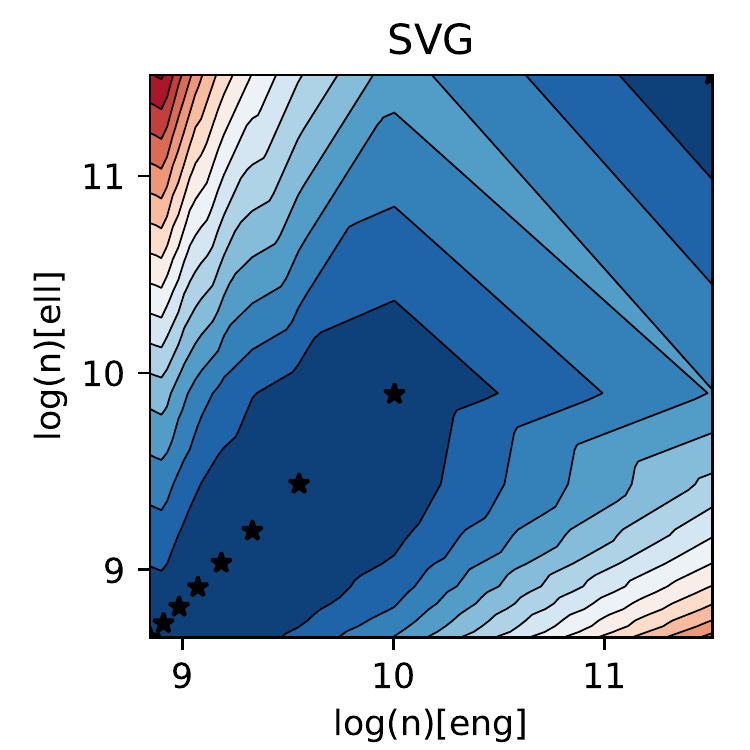}
		\includegraphics[width=\mywidth]{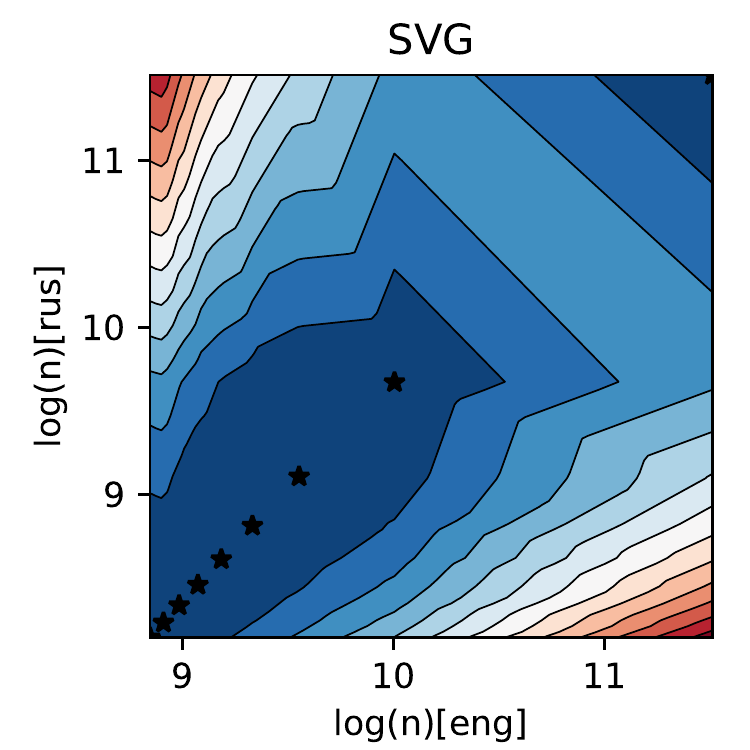}
		\includegraphics[width=\mywidth]{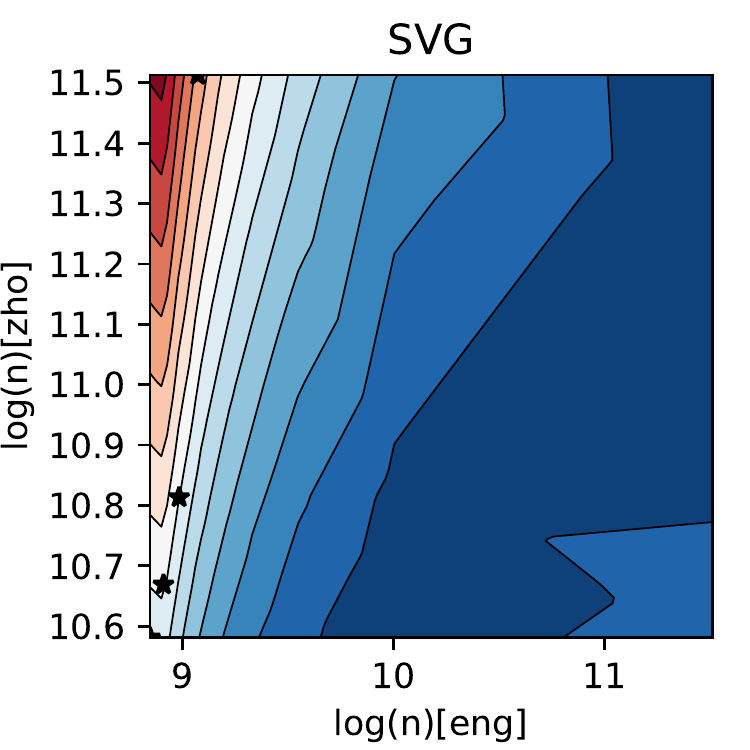}
		\caption{Results showing the interpolated SVG of embedding spaces for different language pairs computed on Wikipedia.  \enote{pd}{todo for myself update and add ell and potentially jpn or korx} \figlabel{spectralwiki}}
	\end{figure}

	\subsection{Contextualized Embedding Learning}
	
	\subsubsection{Fake-English Experiments}
	
	We begin our evaluation with the Fake-English
	experiments. As described in
	\secref{subsec:fakeenglish}, we generate a
	modified ``fake'' English dataset from the English
	Bible corpus. These experiments act as our guide and
	debugging assistant. The behavior of these models can
	tell us a great deal about what to expect in the
	other experiments. Based on these results,
	hyperparameters were tuned both for pretraining and
	for finetuning. Results (on the test set) are shown in the rightmost plot of \figref{bibleres}. In the $EngFake$ experiments, the diagonal is very prominent. Overall, performance in
	these experiments is quite high. On the diagonal, the precision@10 score is the highest in the bottom left corner, while it diminishes slightly towards the top right (detailed results can be found in Table \ref{matrix_en_id} of the appendix). The fact that BERT is able to produce compatible representations between English and Fake-English at different tokenization granularities is in line with findings in \cite{K2020Cross-Lingual}.

	\begin{figure*}
		\centering
		\def\mywidth{0.23\textwidth}
			\centering
			\includegraphics[width=\mywidth]{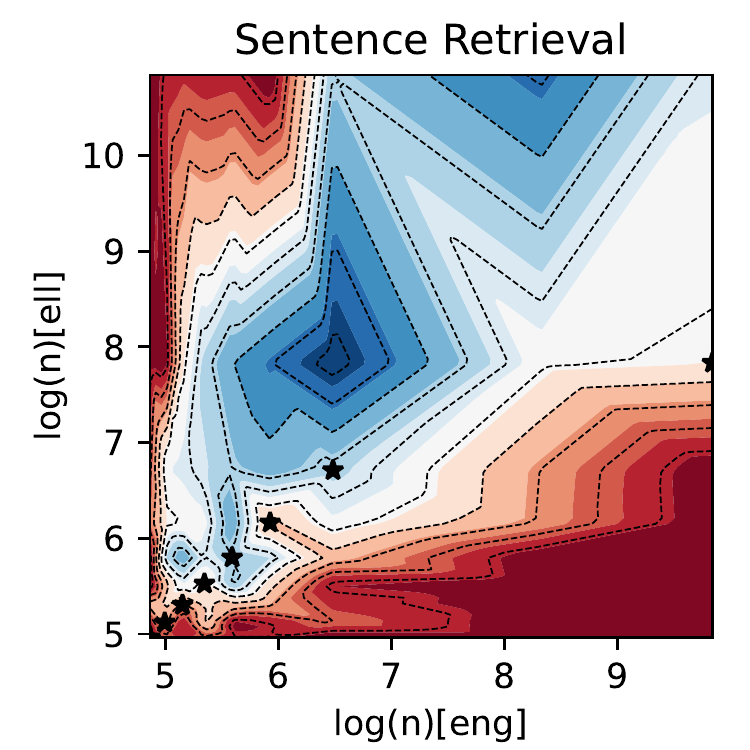}
			\includegraphics[width=\mywidth]{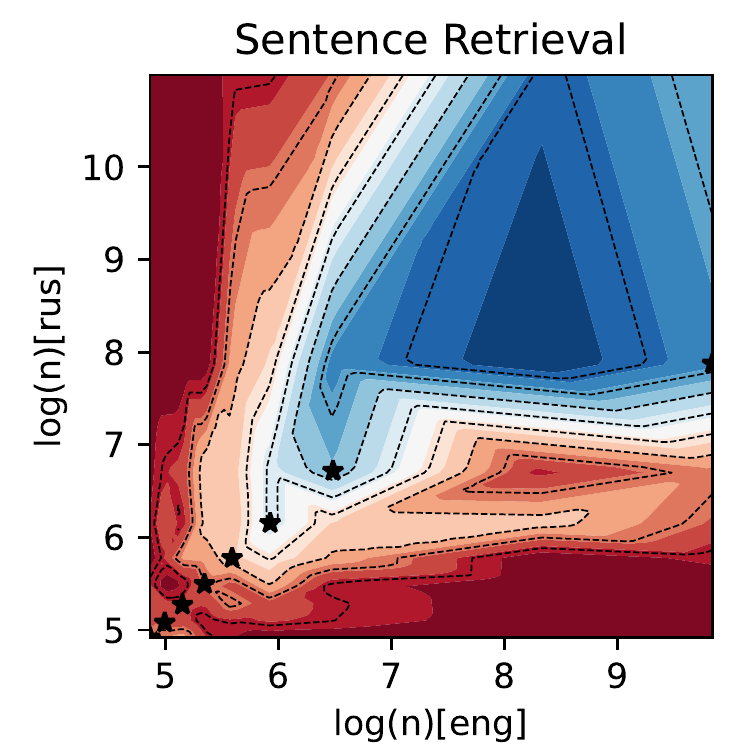}
			\includegraphics[width=\mywidth]{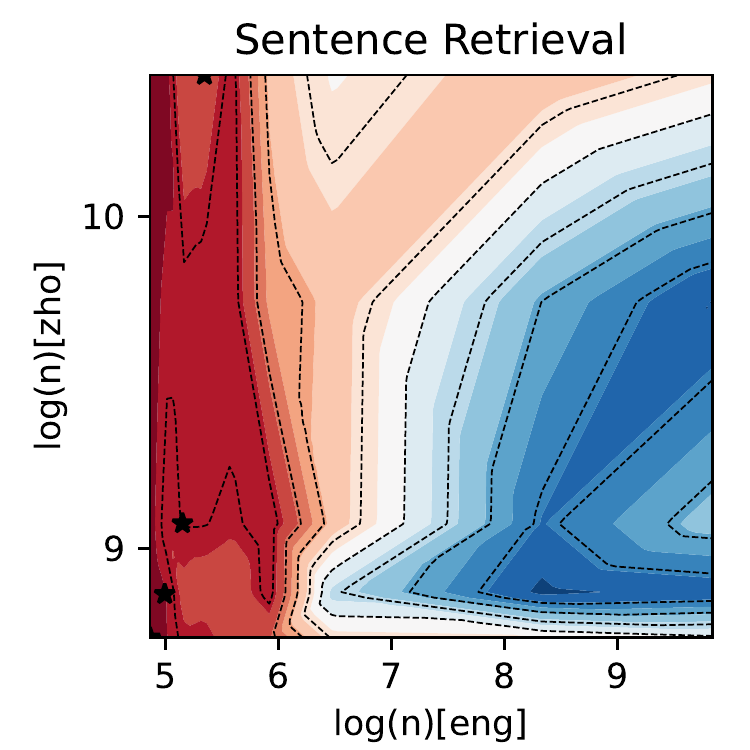}
			\includegraphics[width=\mywidth]{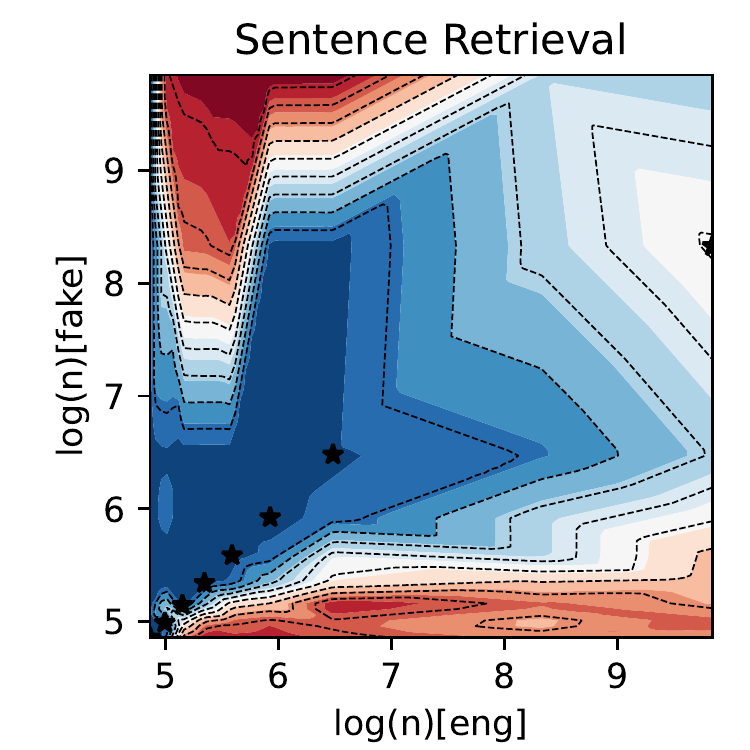}
		\caption{Figures showing the sentence retrieval test results from our Bible experiments. Blue indicates a high sentence retrieval score and red a low score. From left to right: $EngEll$, $EngRus$, $EngZho$ and $EngFake$. \figlabel{bibleres}}
	\end{figure*}

	\subsubsection{Bible Experiments}
	\seclabel{subsub:bibleexps}
	
	As previously described in \secref{subsec:bibleexps}, we conducted experiments on PBC for the $EngEll$, $EngRus$ and $EngZho$ language pairs. Test results are shown in \figref{bibleres} (with detailed tables in Appendix \ref{app:pbc_res_tables}).
	
	$EngEll$ and $EngRus$ both show stronger performance along the diagonal. Performance is low with small sizes (both for English and the paired language). $EngEll$ shows more consistently high precision across the diagonal, from small to large sizes. On the other hand, $EngRus$ performance, while still better along the diagonal, is higher in the top right of the figure where both English and Russian vocabulary sizes are larger.
	
	Despite these similar trends, performance in the $EngRus$ experiment is higher than in $EngEll$ (as indicated by the deep blue in the top right of the figure). Further, in $EngEll$, performance seems to maximize towards the top-middle section of \figref{bibleres}.
	For $EngZho$, instead of a diagonal we see a left-to-right pattern. The larger the English vocabulary size, the better performance is. With the smaller Chinese vocabulary sizes (bottom right of the figure), we see more consistently high results as well. Overall the results show similar trends to findings from the static embeddings in \secref{staticisoresult}.
	
	These results indicate language-specific patterns exist across tokenization sizes, something that can aid us in creating more compatible vocabularies.

	\section{Wikipedia Experiments}
	\seclabel{subsec:wikiexps}
	
	\subsection{Data and Task}
	
	To verify whether our findings generalize to a large-scale setting, we also evaluate on XNLI \cite{conneau2018xnli}. The models we have previously pretrained on the Bible corpus do not have the capacity to tackle this larger task, therefore we opted to instead pretrain new models on Wikipedia data. To this end, English, Greek, Russian and Chinese Wikipedia dumps were used. Due to a lack of computational resources, we kept only 7.4GB of English data, 3.3GB of Chinese and 6.3GB of Russian. For the Greek experiments, all articles were kept resulting in around 0.8GB of data in total. For all other languages, articles were randomly sampled. Details on data size can be found in \tabref{table:wikisizes} of the appendix.
	
	Statistics for XNLI are shown in \tabref{table:xnlistats} of the appendix. Large differences were observed between the two text fields (premise and hypothesis), as well as between Chinese and the other languages (attributed to the difference in scripts).
	
	\subsection{Training}
	\seclabel{subsec:wikitraining}
	
	To accommodate the larger dataset, BERT$_{wiki}$ was used. The number of epochs was set to 1 and the maximum sequence length was reduced to 128 due to computational restrictions.
	Analogously to the small-scale Bible experiments, for pretraining, English data was concatenated with the paired language data, with separate tokenizers trained on the two corpora.
	
	Because evaluating for all possible combinations of size pairs is prohibitively resource-consuming, we settled on three size pairs for each language. These pairs were selected according to the Bible evaluation we ran beforehand. Namely, for each language, we picked the best, the one at the 75th percentile ($perc\_75$) and the worst pair sizes to compare according to their sentence retrieval precision@10 score.
	
	After the pretraining of these models on Wikipedia, we finetuned them on the English XNLI training set and then performed cross-lingual zero-shot evaluation on the other languages. As before, Greek, Russian and Chinese were evaluated.

	\subsection{Evaluation and Results}
	
	\figref{spectralwiki} shows the same experiments as described in \secref{staticisoresult}, this time conducted on the Wikipedia dataset. The key trends are observable here as well. 
	
	In \tabref{table:xnli} we show the accuracy of the
	three examined pairs (as defined
	in \secref{subsec:wikitraining}: best, $perc\_75$, worst) for each language in the XNLI task. Note that a majority baseline has an accuracy of 1/3. The pairs are sorted according to their performance in the Bible experiments (precision@10 on sentence retrieval, largest at the top).
	For all language pairs, performance on the English
	XNLI test set was similar, slightly fluctuating
	($\pm0.75$) around 74\%. These scores are hence
	omitted from \tabref{table:xnli} since they do not
	add much to the discussion.

	The best combinations (from the
	Bible experiments) for $EngEll$ and $EngRus$ have the best XNLI performance as
	well, followed by $perc\_75$ and finally the
	worst-performing combination.
Thus, our findings
for $EngEll$ and $EngRus$ are corroborated: the best performances are found when the absolute compression rate is similar.

	For $EngZho$, we see little variation
 between the three chosen size combinations although the
 best-performing combination is still at the
 top. Maybe this is due to the fact that Chinese behaves
 differently from the two ``alphabetic'' languages and
 therefore a more extensive hyperparameter optimization
 would be required. We leave this question for future research.
 

	\begin{table}[]
		\centering
		\begin{tabular}{|
				>{\columncolor[HTML]{EFEFEF}}c c|}
			\hline
			\cellcolor[HTML]{B5B5B5}$EngEll$ & \cellcolor[HTML]{B5B5B5}Acc. \\ \hline
			$14098_{0.4}$ - $19789_{0.3}$	& 48.8 \\
			$7414_{0.9}$ - $12539_{0.4}$	& 46.4 \\
			$22158_{0.3}$ - $6177_{0.9}$	& 44.6 \\ \hline
			\cellcolor[HTML]{B5B5B5}$EngRus$ & \cellcolor[HTML]{B5B5B5}Acc. \\ \hline
			$22158_{0.3}$ - $15892_{0.3}$	& 59.1 \\
			$9752_{0.6}$ - $15892_{0.3}$	& 58.3 \\
			$6956_{1.0}$ - $15892_{0.3}$	& 56.4 \\ \hline
			\cellcolor[HTML]{B5B5B5}$EngZho$ & \cellcolor[HTML]{B5B5B5}Acc. \\ \hline
			$22158_{0.3}$ - $43005_{0.9}$	& 45.1 \\
			$14098_{0.4}$ - $39393_{1.0}$	& 44.7 \\
			$6956_{1.0}$ - $81426_{0.7}$	& 44.8 \\ \hline
		\end{tabular}
		\caption{Zero-shot XNLI accuracy
		for
English (source) and three target languages. We give
		the vocabulary sizes and, as a subscript,
		Absolute Compression Rate (ACR). 
In each block of three results (one each for ELL, RUS, ZHO),
		the pair of vocabulary sizes with the best performance on 
		sentence retrieval is at the top,
		followed by $perc\_75$ and the worst
		performing pair.
Wikipedia results
 are a good
		predictor of XNLI performance for Greek and Russian.
See text for more details.}
		\tablabel{table:xnli}
	\end{table}

	\subsection{Correlation Analysis}
	
	One major objective is to be able to compare the compatibility of tokenizations across languages. To this end we compute $\log(r^{ABS}_{e})/\log(r^{ABS}_{f})$ for two languages $e$, $f$ and analogously for $r^{REL}$. We compute the correlation of this compatibility measure using the original mBERT tokenizer with the downstream performance of mBERT on XNLI as reported by \cite{hu2020xtreme}. More details can be found in the supplementary. 

	The Pearson correlation is .40 and .34, respectively. This indicates that our measures capture the compatibility of tokenizations and correlate with zero-shot transfer downstream performance. The strength of the correlation is similar to what \citet{rust2020how} find.

	\section{Conclusion}
	
	
	We investigated tokenization compatibility across languages, both for static and contextualized embeddings. To this end, we proposed compression rates and a method to select meaningful vocabulary sizes in an automated manner for any language. We introduced a compatibility measure and showed that it correlates with downstream performance. 
	
	We gave evidence for two key findings that hold for both  static and
	contextualized embeddings. 
	\begin{enumerate*}[label={\textbf{\roman{*})}}]
		\item Tokenization
	compatibility can have a significant impact on
	multilingual performance: performance is generally
	higher when vocabulary sizes are compatible.
	\item Tokenization compatibility varies significantly
	across language pairs: pairs of alphabetic languages show a
	stronger performance on the diagonal (i.e., for
	comparable vocabulary sizes) while
an alphabetic language like English requires large  vocabulary
	sizes when paired with a
        logographic
	language like Chinese.
	\end{enumerate*}
	
	Our study has clear limitations: we
	only experimented with the WordPiece tokenizer and
with a small number of language pairs and there are only a few
	larger-scale experiments on XNLI. We will continue
	research in this direction  and hope that this paper
	will spark interest by other researchers in the compatibility of tokenizations across languages and its effect on multilinguality.

	\section*{Acknowledgements}
	
	This work was supported
	by the European Research Council (\# 740516).
	The second author was supported by the Bavarian research institute for digital transformation (bidt) through their fellowship program. 
	We thank Ehsaneddin Asgari and the anonymous reviewers for valuable comments and fruitful discussions.
	
	\bibliography{anthology,custom}
	\bibliographystyle{acl_natbib}
	
	\clearpage
	\appendix

	\section{Data Statistics}
	
	Here we present details for the PBC and Wikipedia data.
	
	\begin{table}[H]
		\centering
		\begin{tabular}{|c|c|c|c|c|}
			\hline
			\rowcolor[HTML]{B5B5B5} 
			PBC	& Max. & Avg. & Min. & perc\_95  \\
			\cellcolor[HTML]{EFEFEF} ENG	& 474 & 138 & 9 & 249 \\\hline
			\cellcolor[HTML]{EFEFEF} ELL	& 523 & 150 & 3 & 272 \\\hline
			\cellcolor[HTML]{EFEFEF} RUS	& 398 & 122 & 3 & 221 \\\hline
			\cellcolor[HTML]{EFEFEF} ZHO	& 264 & 62 & 8 & 113 \\\hline
		\end{tabular}
		\caption{Character length statistics for the PBC data.}
		\tablabel{table:pbcstats}
	\end{table}

	\begin{table}[H]
		\centering
		\scriptsize
		\begin{tabular}{|c|c|c|c|c|c|c|c|c|}
			\hline
			\rowcolor[HTML]{B5B5B5}
			XNLI	& \multicolumn{2}{|c|}{Max.} & \multicolumn{2}{|c|}{Avg.} & \multicolumn{2}{|c|}{Min.} & \multicolumn{2}{|c|}{perc\_95} \\ \hline
			\cellcolor[HTML]{EFEFEF} ENG	& 1815 & 393 & 113 & 56 & 5 & 1 & 252 & 101 \\\hline
			\cellcolor[HTML]{EFEFEF} ELL	& 350 & 229 & 120 & 59 & 13 & 9 & 225 & 106 \\\hline
			\cellcolor[HTML]{EFEFEF} RUS	& 309 & 195 & 111 & 55 & 11 & 6 & 210 & 100 \\\hline
			\cellcolor[HTML]{EFEFEF} ZHO	& 112 & 87 & 33 & 16 & 4 & 3 & 62 & 29 \\\hline
		\end{tabular}
		\caption{Character length statistics for the XNLI data. We report the lengths of the premise and the hypothesis) separately.}
		\tablabel{table:xnlistats}
	\end{table}
	
	\begin{table}[H]
		\centering
		\begin{tabular}{|c|c|c|c|}
			\hline
			\rowcolor[HTML]{B5B5B5}
			Language	& Train & Dev & Test \\ \hline
			\cellcolor[HTML]{EFEFEF} ENG	& 23,633 & 5,000 & 2,500 \\\hline
			\cellcolor[HTML]{EFEFEF} ELL	& 23,673 & 5,000 & 2,500 \\\hline
			\cellcolor[HTML]{EFEFEF} RUS	& 23,673 & 5,000 & 2,500 \\\hline
			\cellcolor[HTML]{EFEFEF} ZHO	& 23,657 & 5,000 & 2,500 \\\hline
		\end{tabular}
		\caption{Number of verses for each language and split combination in the PBC.}
		\tablabel{table:versesizes}
	\end{table}

	\begin{table}[H]
		\centering
		\begin{tabular}{|
				>{\columncolor[HTML]{EFEFEF}}c |c|}
			\hline
			\cellcolor[HTML]{B5B5B5} Language & \cellcolor[HTML]{B5B5B5} Size \\ \hline
			ENG & 7436MB  \\ \hline
			ELL & 809MB \\ \hline
			RUS & 6291MB \\ \hline
			ZHO & 3252MB \\ \hline
		\end{tabular}
		\caption{Sizes for the languages in our Wikipedia data.}
		\tablabel{table:wikisizes}
	\end{table}

	\section{Computational Details}
	
	For each individual PBC experiment, pretraining took around 2 hours and finetuning/evaluation 25 minutes. For each individual Wikipedia experiment, pretraining took around 145 hours and finetuning/evaluation 5 hours. A multi-GPU server was used, with GeForce GTX 1080Ti devices.
	
	\section{Bible Experiments Detailed Results}
	\label{app:pbc_res_tables}
	
	Here we present the detailed tables for the PBC results. Three runs were made with results averaged. Standard deviations were small, so they were omitted for readability.
	
	\begin{table}[]
		\small
		\setlength\tabcolsep{2.00pt}
		\centering
		\begin{tabular}{c|ccccccccc}
			& 129 & 147 & 172 & 209 & 267 & 374 & 653 & 4139 & 18865\\\hline
			129 & \cellcolor[HTML]{009901}1.00 & \cellcolor[HTML]{009901}0.99 & \cellcolor[HTML]{009901}0.99 & \cellcolor[HTML]{009901}0.99 & \cellcolor[HTML]{009901}0.99 & \cellcolor[HTML]{009901}0.99 & \cellcolor[HTML]{009901}1.00 & \cellcolor[HTML]{009901}1.00 & \cellcolor[HTML]{009901}1.00 \\
			147 & \cellcolor[HTML]{009901}0.96 & \cellcolor[HTML]{009901}0.99 & \cellcolor[HTML]{009901}0.68 & \cellcolor[HTML]{009901}0.97 & \cellcolor[HTML]{009901}0.98 & \cellcolor[HTML]{009901}0.93 & \cellcolor[HTML]{009901}0.97 & \cellcolor[HTML]{009901}0.61 & \cellcolor[HTML]{34FF34}0.11 \\
			172 & \cellcolor[HTML]{32CB00}0.28 & \cellcolor[HTML]{009901}0.53 & \cellcolor[HTML]{009901}1.00 & \cellcolor[HTML]{009901}0.99 & \cellcolor[HTML]{009901}0.99 & \cellcolor[HTML]{009901}0.99 & \cellcolor[HTML]{009901}1.00 & \cellcolor[HTML]{34FF34}0.22 & \cellcolor[HTML]{7FFF7E}0.06 \\
			209 & \cellcolor[HTML]{34FF34}0.11 & \cellcolor[HTML]{32CB00}0.26 & \cellcolor[HTML]{009901}0.71 & \cellcolor[HTML]{009901}0.99 & \cellcolor[HTML]{009901}1.00 & \cellcolor[HTML]{009901}1.00 & \cellcolor[HTML]{009901}1.00 & \cellcolor[HTML]{34FF34}0.21 & \cellcolor[HTML]{7FFF7E}0.05 \\
			267 & \cellcolor[HTML]{34FF34}0.15 & \cellcolor[HTML]{34FF34}0.20 & \cellcolor[HTML]{32CB00}0.44 & \cellcolor[HTML]{009901}0.94 & \cellcolor[HTML]{009901}1.00 & \cellcolor[HTML]{009901}0.99 & \cellcolor[HTML]{009901}1.00 & \cellcolor[HTML]{34FF34}0.16 & \cellcolor[HTML]{7FFF7E}0.05 \\
			374 & \cellcolor[HTML]{34FF34}0.13 & \cellcolor[HTML]{34FF34}0.17 & \cellcolor[HTML]{32CB00}0.37 & \cellcolor[HTML]{009901}0.72 & \cellcolor[HTML]{009901}0.94 & \cellcolor[HTML]{009901}0.98 & \cellcolor[HTML]{009901}1.00 & \cellcolor[HTML]{009901}0.99 & \cellcolor[HTML]{BDFFBC}0.03 \\
			653 & \cellcolor[HTML]{34FF34}0.18 & \cellcolor[HTML]{34FF34}0.22 & \cellcolor[HTML]{34FF34}0.10 & \cellcolor[HTML]{32CB00}0.47 & \cellcolor[HTML]{009901}0.57 & \cellcolor[HTML]{009901}0.94 & \cellcolor[HTML]{009901}0.97 & \cellcolor[HTML]{009901}0.99 & \cellcolor[HTML]{BDFFBC}0.03 \\
			4139 & \cellcolor[HTML]{32CB00}0.26 & \cellcolor[HTML]{32CB00}0.35 & \cellcolor[HTML]{34FF34}0.22 & \cellcolor[HTML]{32CB00}0.39 & \cellcolor[HTML]{009901}0.66 & \cellcolor[HTML]{009901}0.66 & \cellcolor[HTML]{009901}0.89 & \cellcolor[HTML]{009901}0.67 & \cellcolor[HTML]{009901}0.65 \\
			18865 & \cellcolor[HTML]{32CB00}0.28 & \cellcolor[HTML]{34FF34}0.21 & \cellcolor[HTML]{32CB00}0.32 & \cellcolor[HTML]{32CB00}0.36 & \cellcolor[HTML]{32CB00}0.37 & \cellcolor[HTML]{009901}0.51 & \cellcolor[HTML]{009901}0.69 & \cellcolor[HTML]{32CB00}0.49 & \cellcolor[HTML]{009901}0.68 \\
		\end{tabular}
		\caption{Similarity Matrix for $EngFake$. Rows denote English and columns Fake-English.}
		\label{matrix_en_id}
	\end{table}

	\begin{table}[]
	\small
	\setlength\tabcolsep{2.00pt}
	\centering
	\begin{tabular}{c|ccccccccc}
		& 143 & 167 & 201 & 251 & 330 & 475 & 821 & 2524 & 51077\\\hline
		129 & \cellcolor[HTML]{BDFFBC}0.00 & \cellcolor[HTML]{7FFF7E}0.03 & \cellcolor[HTML]{34FF34}0.09 & \cellcolor[HTML]{BDFFBC}0.01 & \cellcolor[HTML]{BDFFBC}0.01 & \cellcolor[HTML]{34FF34}0.06 & \cellcolor[HTML]{7FFF7E}0.04 & \cellcolor[HTML]{7FFF7E}0.02 & \cellcolor[HTML]{BDFFBC}0.00 \\
		147 & \cellcolor[HTML]{34FF34}0.05 & \cellcolor[HTML]{34FF34}0.09 & \cellcolor[HTML]{32CB00}0.11 & \cellcolor[HTML]{34FF34}0.09 & \cellcolor[HTML]{32CB00}0.16 & \cellcolor[HTML]{32CB00}0.12 & \cellcolor[HTML]{32CB00}0.14 & \cellcolor[HTML]{BDFFBC}0.00 & \cellcolor[HTML]{7FFF7E}0.03 \\
		172 & \cellcolor[HTML]{7FFF7E}0.04 & \cellcolor[HTML]{7FFF7E}0.02 & \cellcolor[HTML]{34FF34}0.07 & \cellcolor[HTML]{32CB00}0.14 & \cellcolor[HTML]{009901}0.20 & \cellcolor[HTML]{32CB00}0.13 & \cellcolor[HTML]{32CB00}0.14 & \cellcolor[HTML]{32CB00}0.12 & \cellcolor[HTML]{7FFF7E}0.04 \\
		209 & \cellcolor[HTML]{34FF34}0.05 & \cellcolor[HTML]{32CB00}0.10 & \cellcolor[HTML]{32CB00}0.10 & \cellcolor[HTML]{32CB00}0.12 & \cellcolor[HTML]{32CB00}0.15 & \cellcolor[HTML]{32CB00}0.13 & \cellcolor[HTML]{32CB00}0.16 & \cellcolor[HTML]{32CB00}0.17 & \cellcolor[HTML]{7FFF7E}0.03 \\
		267 & \cellcolor[HTML]{7FFF7E}0.03 & \cellcolor[HTML]{BDFFBC}0.01 & \cellcolor[HTML]{34FF34}0.09 & \cellcolor[HTML]{32CB00}0.18 & \cellcolor[HTML]{32CB00}0.17 & \cellcolor[HTML]{32CB00}0.19 & \cellcolor[HTML]{32CB00}0.18 & \cellcolor[HTML]{009901}0.20 & \cellcolor[HTML]{7FFF7E}0.03 \\
		374 & \cellcolor[HTML]{7FFF7E}0.03 & \cellcolor[HTML]{7FFF7E}0.02 & \cellcolor[HTML]{34FF34}0.08 & \cellcolor[HTML]{32CB00}0.11 & \cellcolor[HTML]{32CB00}0.16 & \cellcolor[HTML]{34FF34}0.09 & \cellcolor[HTML]{32CB00}0.19 & \cellcolor[HTML]{009901}0.22 & \cellcolor[HTML]{BDFFBC}0.00 \\
		653 & \cellcolor[HTML]{BDFFBC}0.00 & \cellcolor[HTML]{34FF34}0.05 & \cellcolor[HTML]{7FFF7E}0.04 & \cellcolor[HTML]{7FFF7E}0.02 & \cellcolor[HTML]{34FF34}0.09 & \cellcolor[HTML]{32CB00}0.14 & \cellcolor[HTML]{32CB00}0.17 & \cellcolor[HTML]{009901}0.26 & \cellcolor[HTML]{32CB00}0.18 \\
		4139 & \cellcolor[HTML]{BDFFBC}0.00 & \cellcolor[HTML]{BDFFBC}0.00 & \cellcolor[HTML]{BDFFBC}0.00 & \cellcolor[HTML]{BDFFBC}0.01 & \cellcolor[HTML]{BDFFBC}0.00 & \cellcolor[HTML]{34FF34}0.07 & \cellcolor[HTML]{34FF34}0.07 & \cellcolor[HTML]{32CB00}0.13 & \cellcolor[HTML]{009901}0.23 \\
		18865 & \cellcolor[HTML]{BDFFBC}0.01 & \cellcolor[HTML]{BDFFBC}0.00 & \cellcolor[HTML]{BDFFBC}0.00 & \cellcolor[HTML]{BDFFBC}0.00 & \cellcolor[HTML]{BDFFBC}0.00 & \cellcolor[HTML]{BDFFBC}0.01 & \cellcolor[HTML]{BDFFBC}0.00 & \cellcolor[HTML]{32CB00}0.12 & \cellcolor[HTML]{32CB00}0.14 \\
	\end{tabular}
	\caption{Similarity Matrix for $EngEll$. Rows denote English and columns Greek.}
	\label{matrix_en_el}
	\end{table}

	\begin{table}[]
		\small
		\setlength\tabcolsep{2.00pt}
		\centering
		\begin{tabular}{c|ccccccccc}
			& 137 & 161 & 195 & 244 & 323 & 471 & 827 & 2632 & 59341\\\hline
			129  & \cellcolor[HTML]{34FF34}0.09 & \cellcolor[HTML]{34FF34}0.09 & \cellcolor[HTML]{34FF34}0.07 & \cellcolor[HTML]{34FF34}0.06 & \cellcolor[HTML]{BDFFBC}0.01 & \cellcolor[HTML]{7FFF7E}0.03 & \cellcolor[HTML]{7FFF7E}0.02 & \cellcolor[HTML]{7FFF7E}0.02 & \cellcolor[HTML]{BDFFBC}0.01 \\
			147  & \cellcolor[HTML]{32CB00}0.12 & \cellcolor[HTML]{34FF34}0.08 & \cellcolor[HTML]{34FF34}0.07 & \cellcolor[HTML]{BDFFBC}0.01 & \cellcolor[HTML]{34FF34}0.06 & \cellcolor[HTML]{32CB00}0.10 & \cellcolor[HTML]{34FF34}0.06 & \cellcolor[HTML]{BDFFBC}0.00 & \cellcolor[HTML]{BDFFBC}0.00 \\
			172  & \cellcolor[HTML]{32CB00}0.14 & \cellcolor[HTML]{34FF34}0.08 & \cellcolor[HTML]{34FF34}0.07 & \cellcolor[HTML]{7FFF7E}0.03 & \cellcolor[HTML]{32CB00}0.13 & \cellcolor[HTML]{7FFF7E}0.04 & \cellcolor[HTML]{34FF34}0.07 & \cellcolor[HTML]{BDFFBC}0.00 & \cellcolor[HTML]{BDFFBC}0.00 \\
			209  & \cellcolor[HTML]{34FF34}0.06 & \cellcolor[HTML]{7FFF7E}0.03 & \cellcolor[HTML]{34FF34}0.06 & \cellcolor[HTML]{32CB00}0.10 & \cellcolor[HTML]{32CB00}0.13 & \cellcolor[HTML]{32CB00}0.16 & \cellcolor[HTML]{32CB00}0.17 & \cellcolor[HTML]{BDFFBC}0.00 & \cellcolor[HTML]{BDFFBC}0.01 \\
			267  & \cellcolor[HTML]{7FFF7E}0.03 & \cellcolor[HTML]{34FF34}0.05 & \cellcolor[HTML]{32CB00}0.11 & \cellcolor[HTML]{34FF34}0.08 & \cellcolor[HTML]{32CB00}0.17 & \cellcolor[HTML]{32CB00}0.16 & \cellcolor[HTML]{32CB00}0.16 & \cellcolor[HTML]{32CB00}0.13 & \cellcolor[HTML]{7FFF7E}0.04 \\
			374  & \cellcolor[HTML]{BDFFBC}0.01 & \cellcolor[HTML]{34FF34}0.06 & \cellcolor[HTML]{34FF34}0.08 & \cellcolor[HTML]{32CB00}0.15 & \cellcolor[HTML]{009901}0.20 & \cellcolor[HTML]{009901}0.26 & \cellcolor[HTML]{009901}0.26 & \cellcolor[HTML]{32CB00}0.19 & \cellcolor[HTML]{34FF34}0.05 \\
			653  & \cellcolor[HTML]{BDFFBC}0.00 & \cellcolor[HTML]{34FF34}0.05 & \cellcolor[HTML]{34FF34}0.05 & \cellcolor[HTML]{7FFF7E}0.04 & \cellcolor[HTML]{32CB00}0.14 & \cellcolor[HTML]{32CB00}0.18 & \cellcolor[HTML]{009901}0.33 & \cellcolor[HTML]{009901}0.37 & \cellcolor[HTML]{34FF34}0.09 \\
			4139  & \cellcolor[HTML]{BDFFBC}0.00 & \cellcolor[HTML]{BDFFBC}0.00 & \cellcolor[HTML]{BDFFBC}0.00 & \cellcolor[HTML]{BDFFBC}0.00 & \cellcolor[HTML]{BDFFBC}0.00 & \cellcolor[HTML]{32CB00}0.17 & \cellcolor[HTML]{34FF34}0.05 & \cellcolor[HTML]{009901}0.44 & \cellcolor[HTML]{009901}0.41 \\
			18865  & \cellcolor[HTML]{BDFFBC}0.00 & \cellcolor[HTML]{BDFFBC}0.00 & \cellcolor[HTML]{BDFFBC}0.00 & \cellcolor[HTML]{BDFFBC}0.00 & \cellcolor[HTML]{7FFF7E}0.04 & \cellcolor[HTML]{34FF34}0.09 & \cellcolor[HTML]{32CB00}0.11 & \cellcolor[HTML]{009901}0.37 & \cellcolor[HTML]{009901}0.33 \\
		\end{tabular}
		\caption{Similarity Matrix for $EngRuS$. Rows denote English and columns Russian.}
		\label{matrix_en_ru}
	\end{table}
	
	\begin{table}[]
		\small
		\setlength\tabcolsep{2.00pt}
		\centering
		\begin{tabular}{c|ccccc}
			& 6189 & 7054 & 8733 & 17083 & 33786\\\hline
			129 & \cellcolor[HTML]{BDFFBC}0.00 & \cellcolor[HTML]{BDFFBC}0.01 & \cellcolor[HTML]{BDFFBC}0.01 & \cellcolor[HTML]{BDFFBC}0.00 & \cellcolor[HTML]{BDFFBC}0.01 \\
			147 & \cellcolor[HTML]{7FFF7E}0.02 & \cellcolor[HTML]{7FFF7E}0.02 & \cellcolor[HTML]{34FF34}0.05 & \cellcolor[HTML]{7FFF7E}0.03 & \cellcolor[HTML]{BDFFBC}0.01 \\
			172 & \cellcolor[HTML]{34FF34}0.05 & \cellcolor[HTML]{34FF34}0.06 & \cellcolor[HTML]{7FFF7E}0.04 & \cellcolor[HTML]{7FFF7E}0.03 & \cellcolor[HTML]{34FF34}0.07 \\
			209 & \cellcolor[HTML]{7FFF7E}0.04 & \cellcolor[HTML]{34FF34}0.06 & \cellcolor[HTML]{7FFF7E}0.04 & \cellcolor[HTML]{7FFF7E}0.03 & \cellcolor[HTML]{34FF34}0.07 \\
			267 & \cellcolor[HTML]{34FF34}0.06 & \cellcolor[HTML]{34FF34}0.07 & \cellcolor[HTML]{34FF34}0.05 & \cellcolor[HTML]{7FFF7E}0.03 & \cellcolor[HTML]{7FFF7E}0.03 \\
			374 & \cellcolor[HTML]{34FF34}0.07 & \cellcolor[HTML]{7FFF7E}0.03 & \cellcolor[HTML]{7FFF7E}0.03 & \cellcolor[HTML]{32CB00}0.11 & \cellcolor[HTML]{32CB00}0.13 \\
			653 & \cellcolor[HTML]{32CB00}0.16 & \cellcolor[HTML]{009901}0.24 & \cellcolor[HTML]{32CB00}0.13 & \cellcolor[HTML]{32CB00}0.13 & \cellcolor[HTML]{32CB00}0.18 \\
			4139 & \cellcolor[HTML]{32CB00}0.17 & \cellcolor[HTML]{009901}0.36 & \cellcolor[HTML]{009901}0.33 & \cellcolor[HTML]{009901}0.28 & \cellcolor[HTML]{32CB00}0.13 \\
			18865 & \cellcolor[HTML]{32CB00}0.19 & \cellcolor[HTML]{009901}0.34 & \cellcolor[HTML]{009901}0.26 & \cellcolor[HTML]{009901}0.35 & \cellcolor[HTML]{32CB00}0.17 \\
		\end{tabular}
		\caption{Similarity Matrix for $EngZho$. Rows denote English and columns Chinese.}
		\label{matrix_en_zh}
	\end{table}

	\section{Correlation Analysis}
	
	We show detailed results for the correlation analysis in \tabref{correlation}.
	
	\begin{table*}[t]
\centering
\footnotesize
	\begin{tabular}{c|c|cc|cc}
		Language & XNLI Acc. & $r^{ABS}$ & $r^{REL}$ & $\log(r^{ABS}_{x})/\log(r^{ABS}_{eng})$ & $\log(r^{REL}_{x})/\log(r^{REL}_{eng})$ \\
		\midrule
		ara & 64.9 & 0.46 & 0.06 & 0.64 & 0.68 \\
		bul & 68.9 & 0.39 & 0.03 & 0.77 & 0.85 \\
		deu & 71.1 & 0.31 & 0.01 & 0.95 & 1.06 \\
		ell & 66.4 & 0.57 & 0.06 & 0.46 & 0.65 \\
		eng & 81.4 & 0.30 & 0.01 & 1.00 & 1.00 \\
		esp & 74.3 & 0.32 & 0.02 & 0.95 & 0.93 \\
		fra & 73.8 & 0.33 & 0.02 & 0.91 & 0.95 \\
		hin & 60.0 & 0.53 & 0.08 & 0.53 & 0.61 \\
		rus & 69.0 & 0.37 & 0.02 & 0.82 & 0.90 \\
		swa & 50.4 & 0.39 & 0.03 & 0.78 & 0.84 \\
		tha & 55.8 & 0.68 & 0.12 & 0.31 & 0.49 \\
		tur & 61.6 & 0.37 & 0.03 & 0.81 & 0.81 \\
		urd & 58.0 & 0.48 & 0.07 & 0.60 & 0.63 \\
		vie & 69.5 & 0.45 & 0.08 & 0.65 & 0.60 \\
		zho & 69.3 & 0.96 & 0.69 & 0.04 & 0.09 \\
		\midrule
		\midrule
		Pearson Correlation& & & & 0.40 & 0.34 \\
	\end{tabular}
\caption{Compression rates computed using the mBERT tokenizer. XNLI results by \citet{hu2020xtreme}.}\tablabel{correlation}
	\end{table*}
	
	\section{SVG Plots}
	\label{app:svg_plots}
	
	In the next pages, we show the complete SVG plots for all 61 languages.
	
	\begin{figure*}
		\centering
		\def\mywidth{0.24\textwidth}
		\centering
		\includegraphics[width=\mywidth]{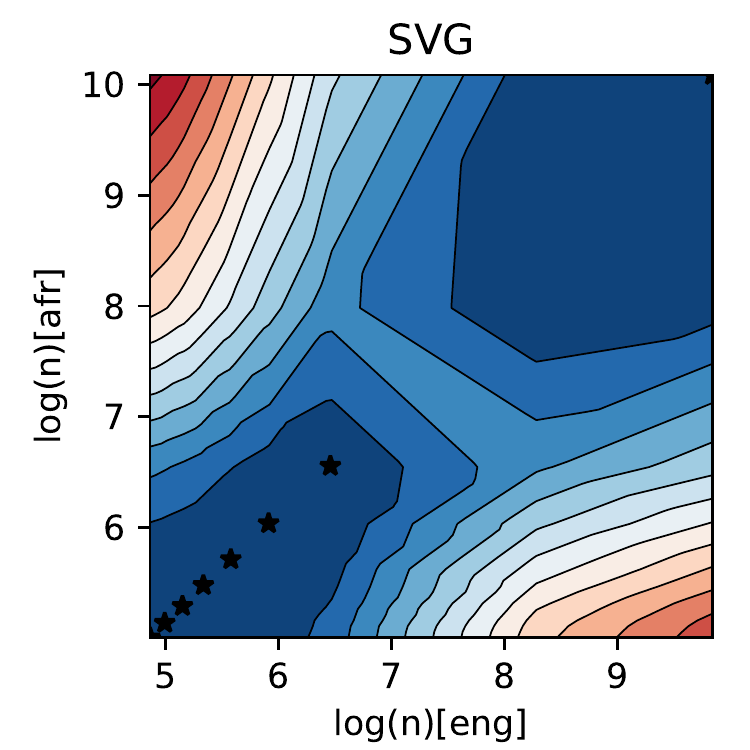}
		\includegraphics[width=\mywidth]{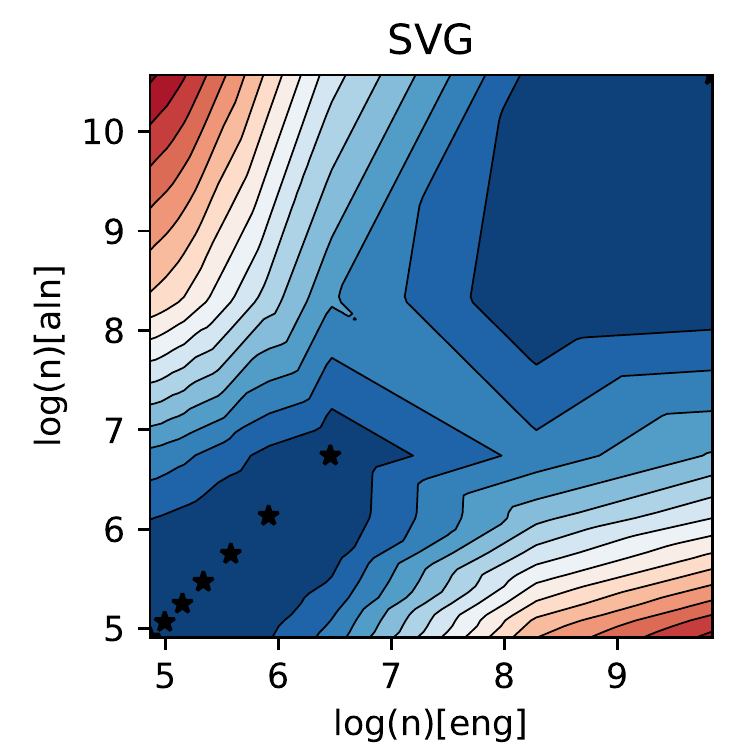}
		\includegraphics[width=\mywidth]{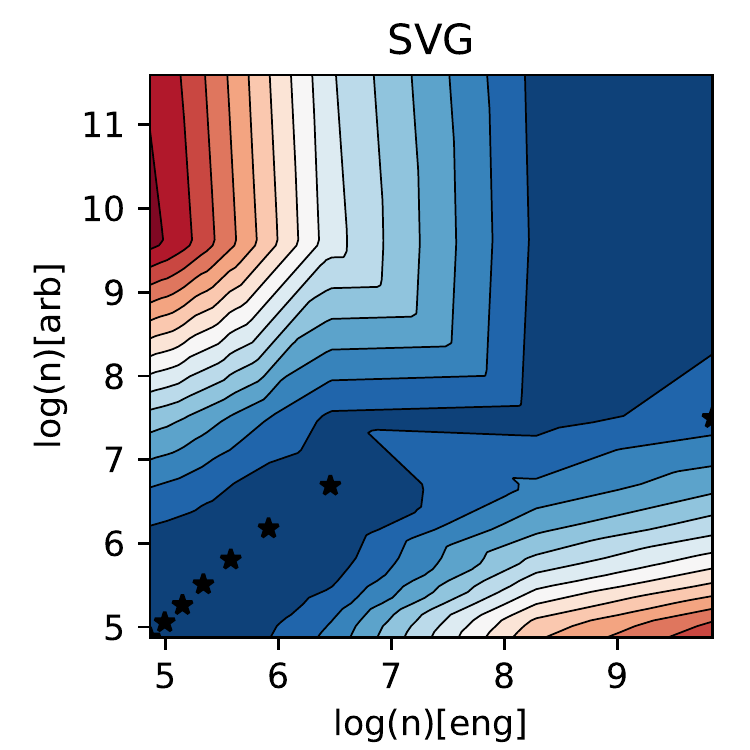}
		\includegraphics[width=\mywidth]{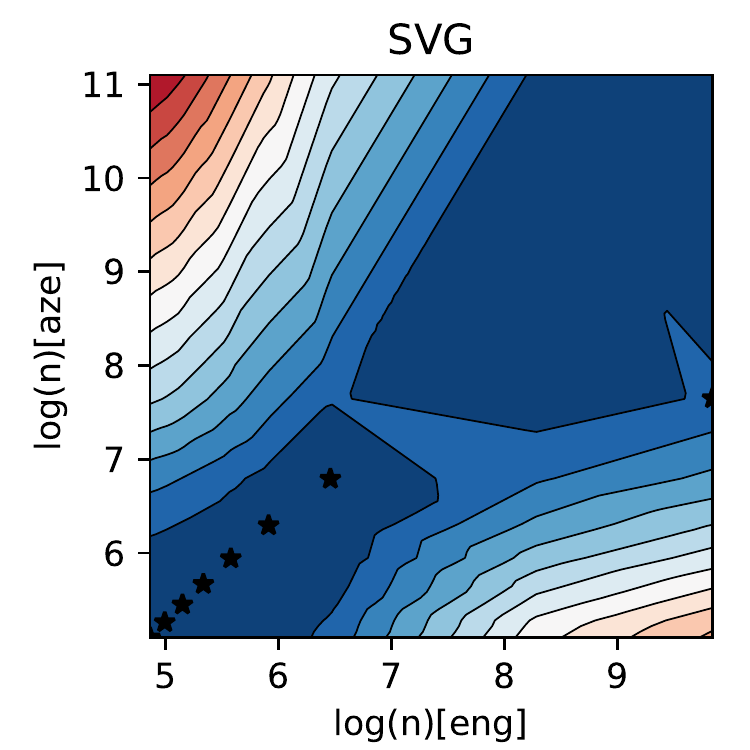}
		\includegraphics[width=\mywidth]{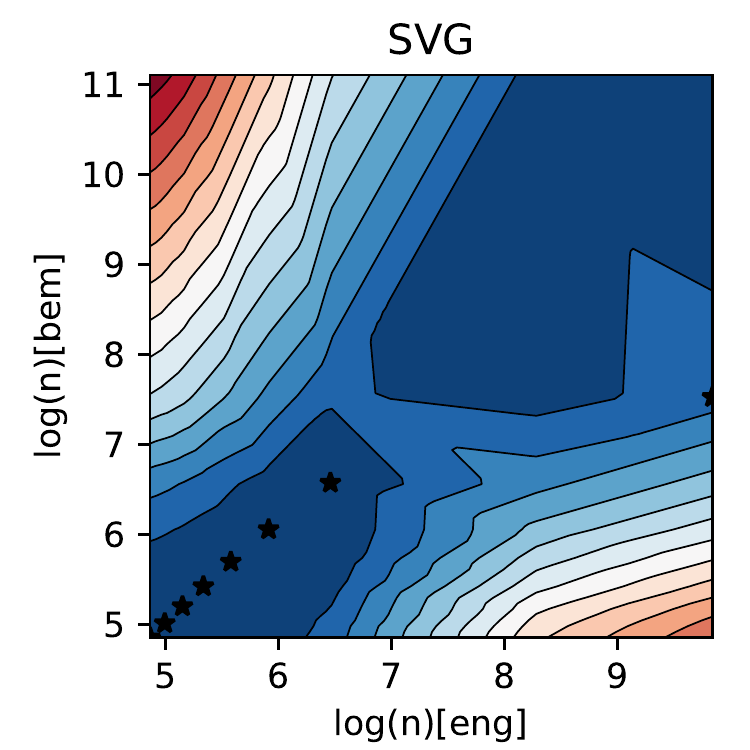}
		\includegraphics[width=\mywidth]{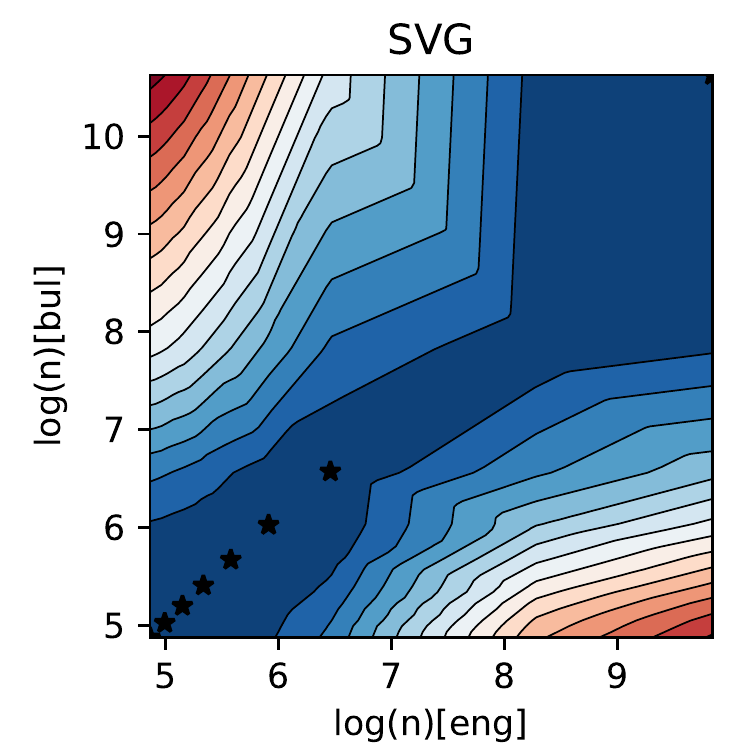}
		\includegraphics[width=\mywidth]{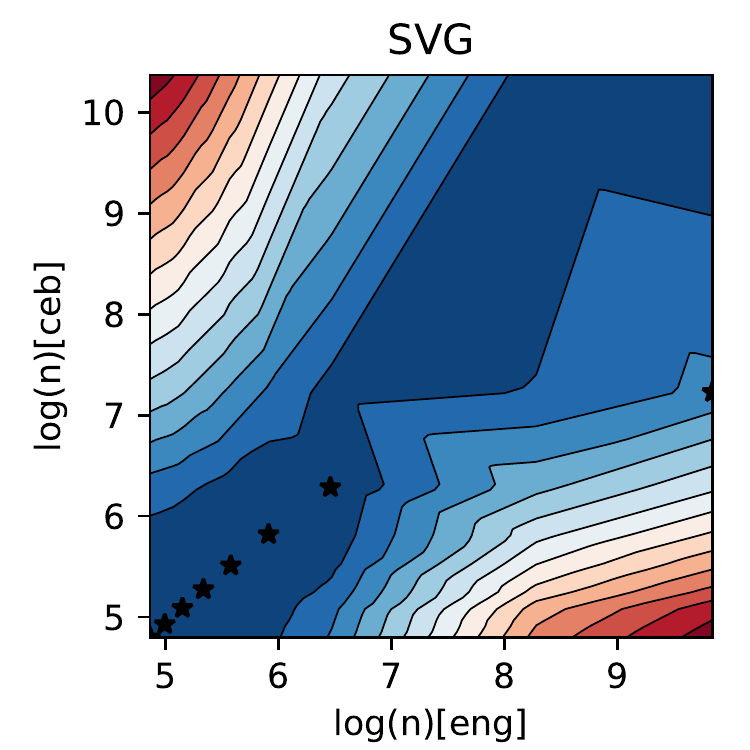}
		\includegraphics[width=\mywidth]{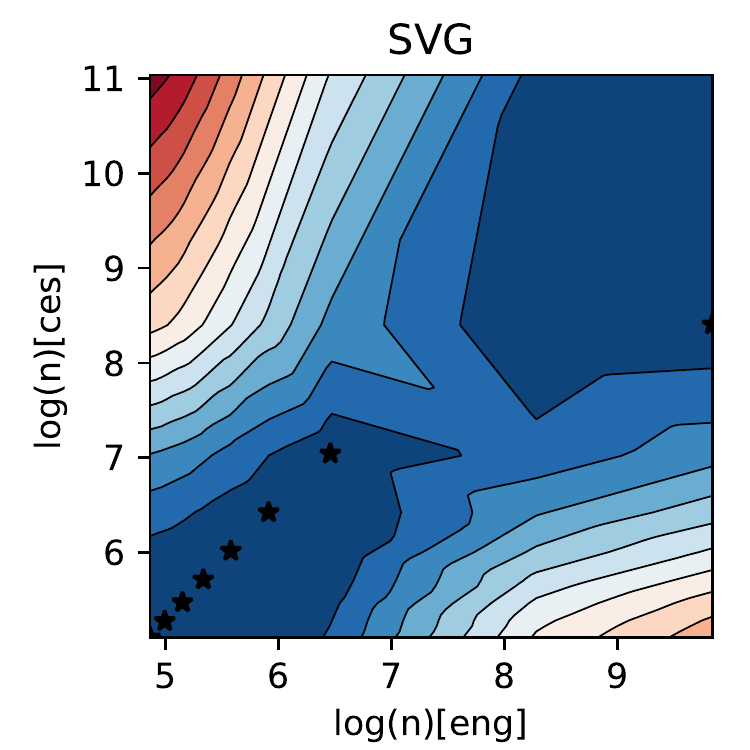}
		\includegraphics[width=\mywidth]{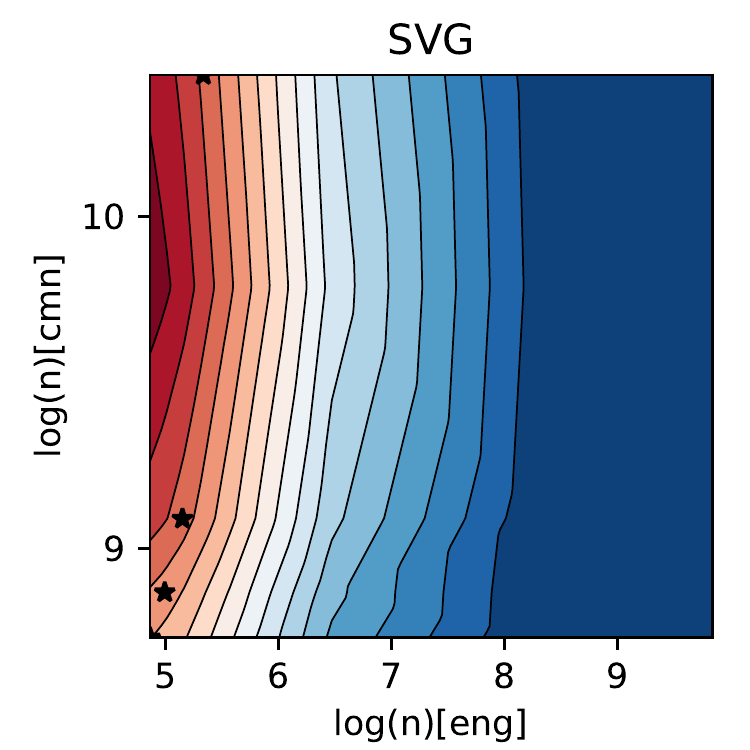}
		\includegraphics[width=\mywidth]{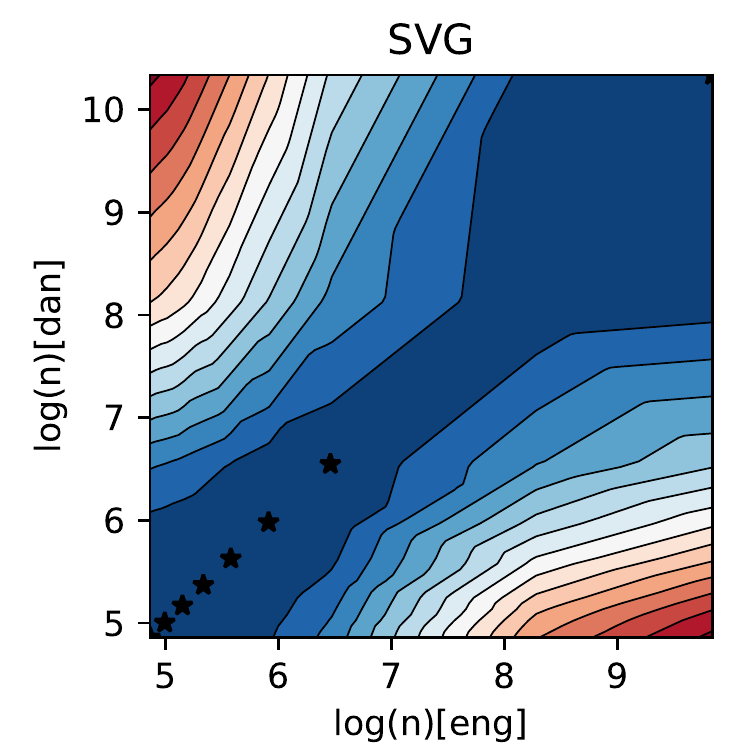}
		\includegraphics[width=\mywidth]{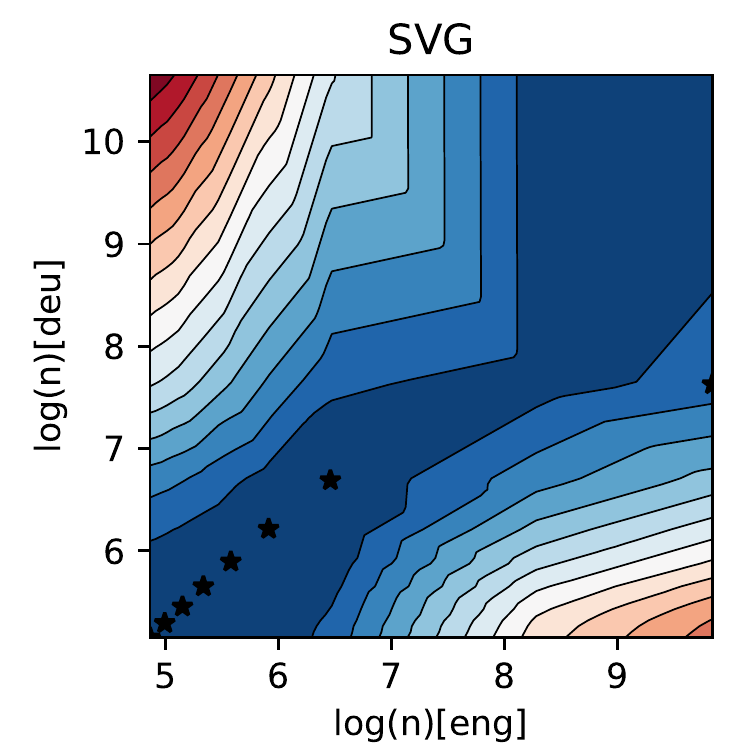}
		\includegraphics[width=\mywidth]{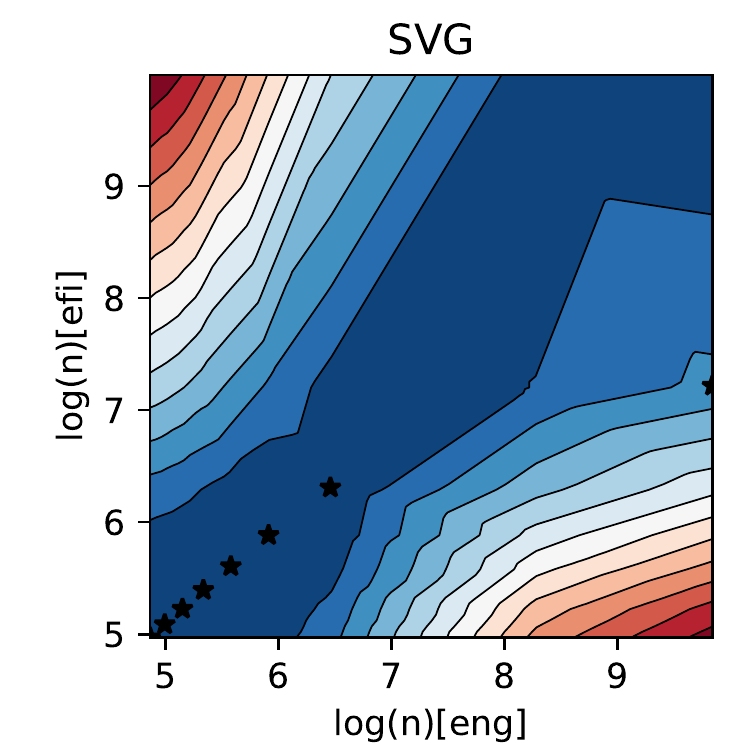}
		\includegraphics[width=\mywidth]{assets/plots3,12,svg/12,eng_newworld2013,ell_newworld}
		\includegraphics[width=\mywidth]{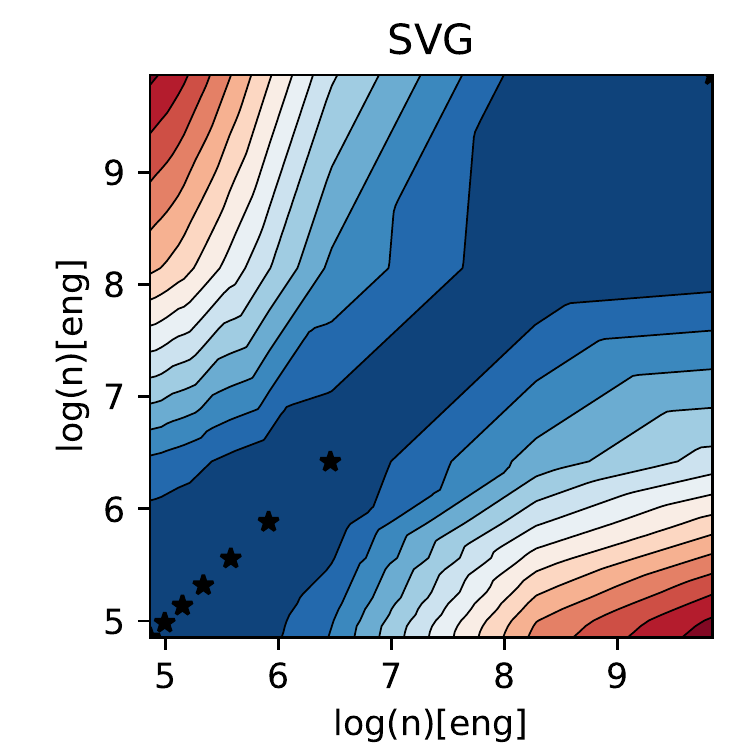}
		\includegraphics[width=\mywidth]{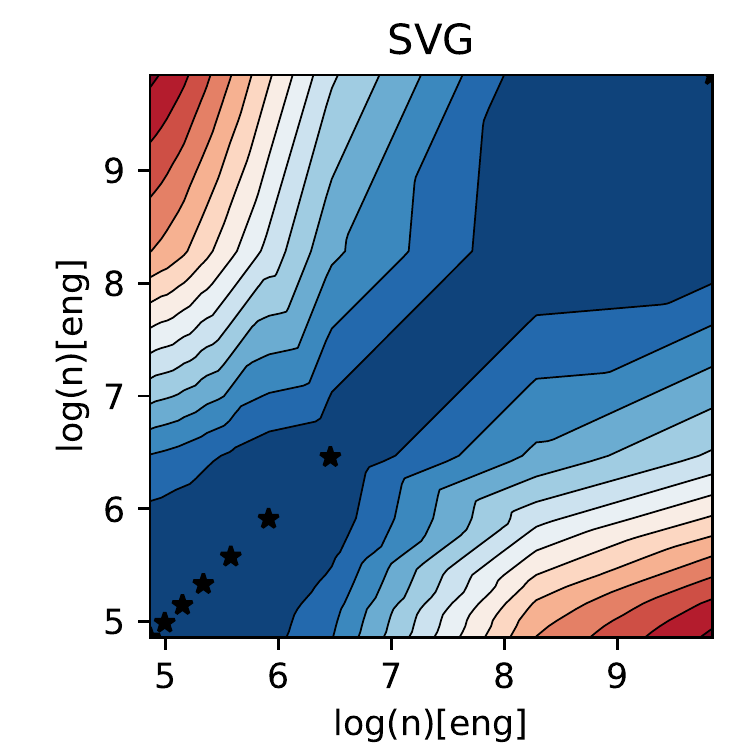}
		\includegraphics[width=\mywidth]{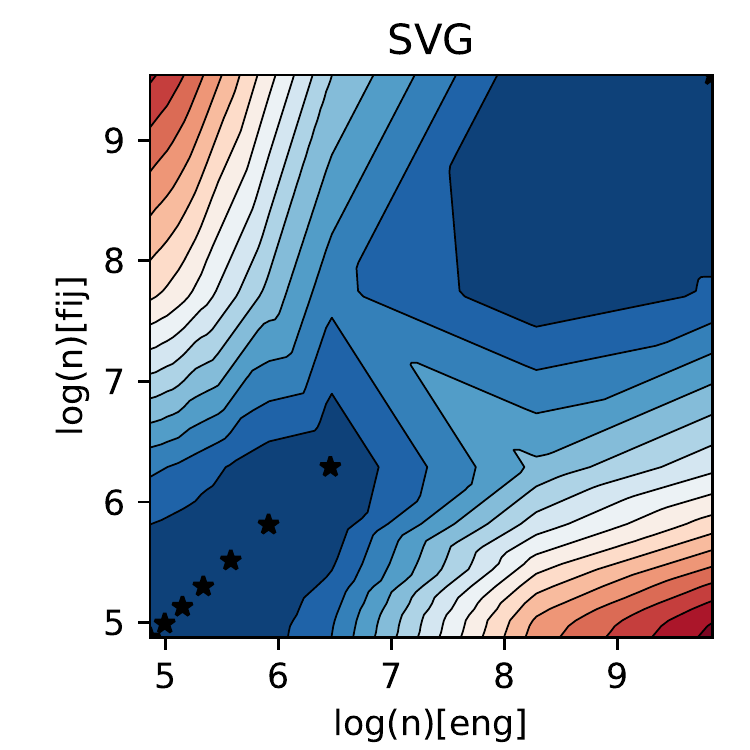}
		\includegraphics[width=\mywidth]{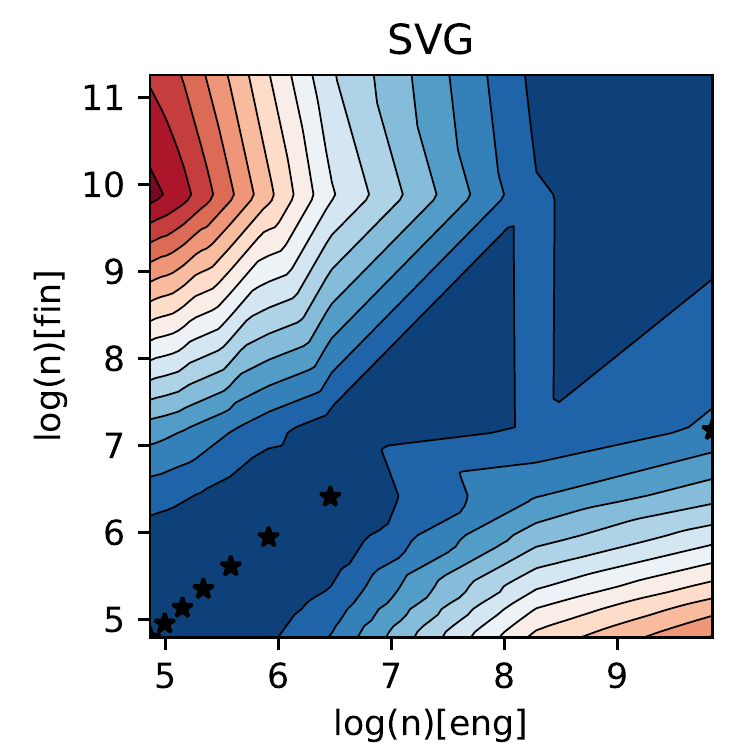}
		\includegraphics[width=\mywidth]{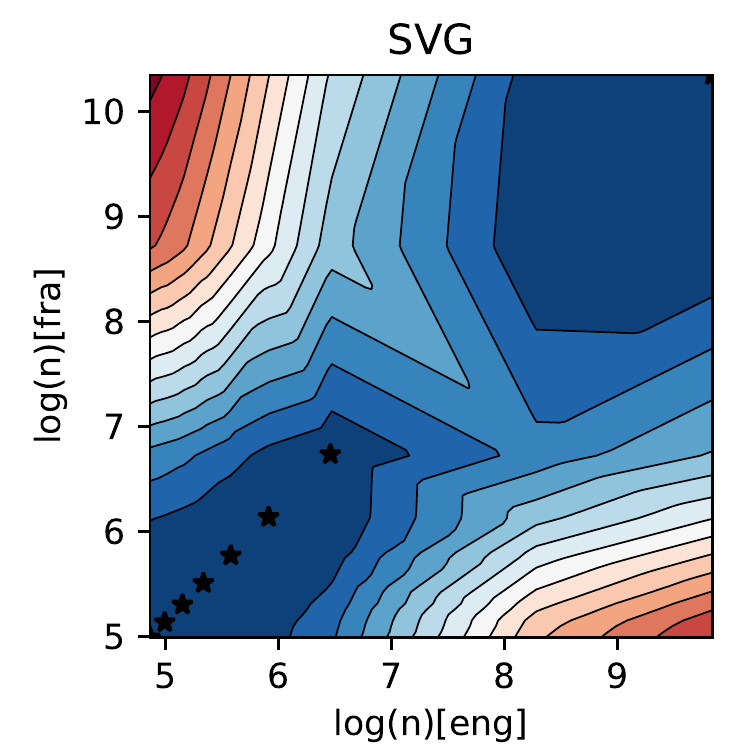}
		\includegraphics[width=\mywidth]{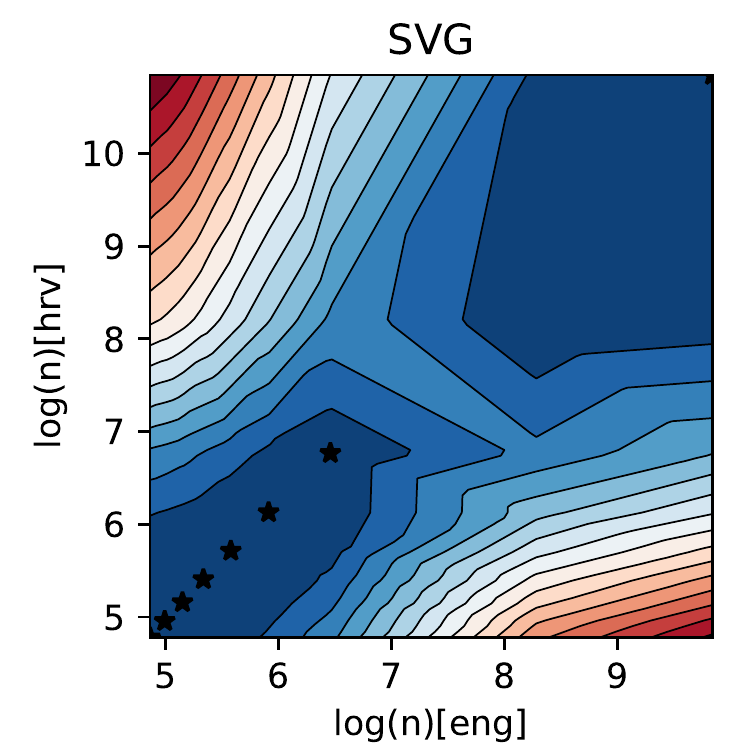}
		\includegraphics[width=\mywidth]{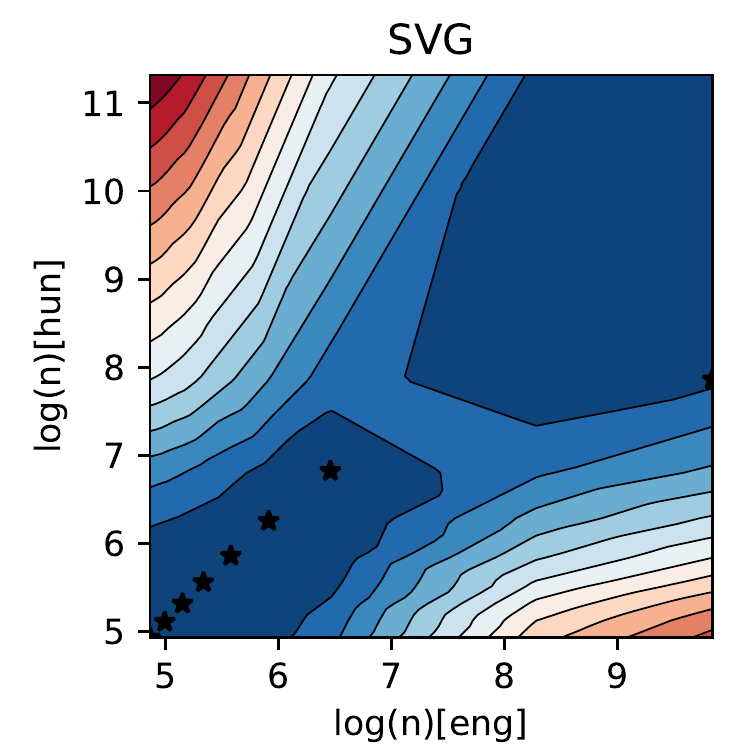}
		\caption{Figures showing the interpolated SVG for different language pairs as computed on the PBC (1/4).}
	\end{figure*}
	
	\begin{figure*}
		\centering
		\def\mywidth{0.24\textwidth}
		\centering
		\includegraphics[width=\mywidth]{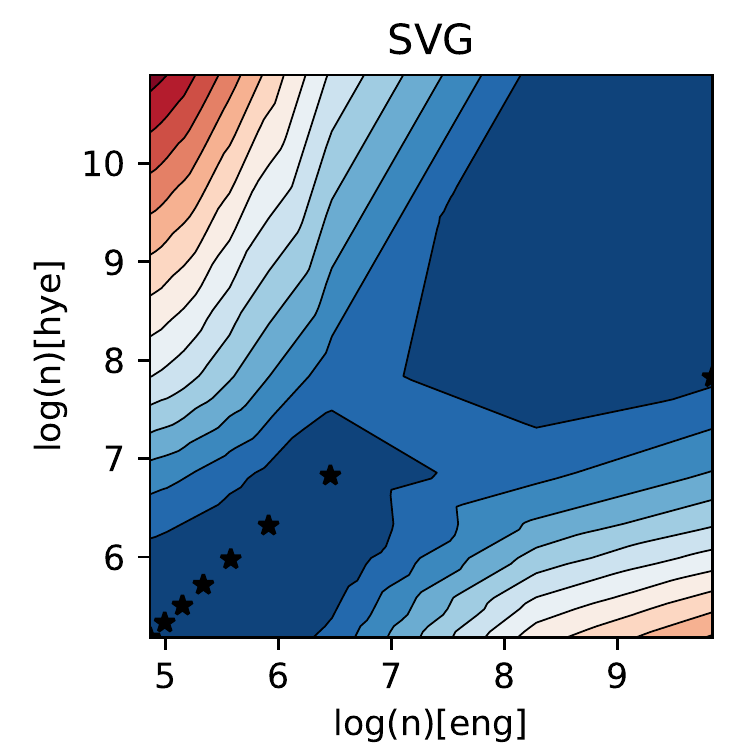}
		\includegraphics[width=\mywidth]{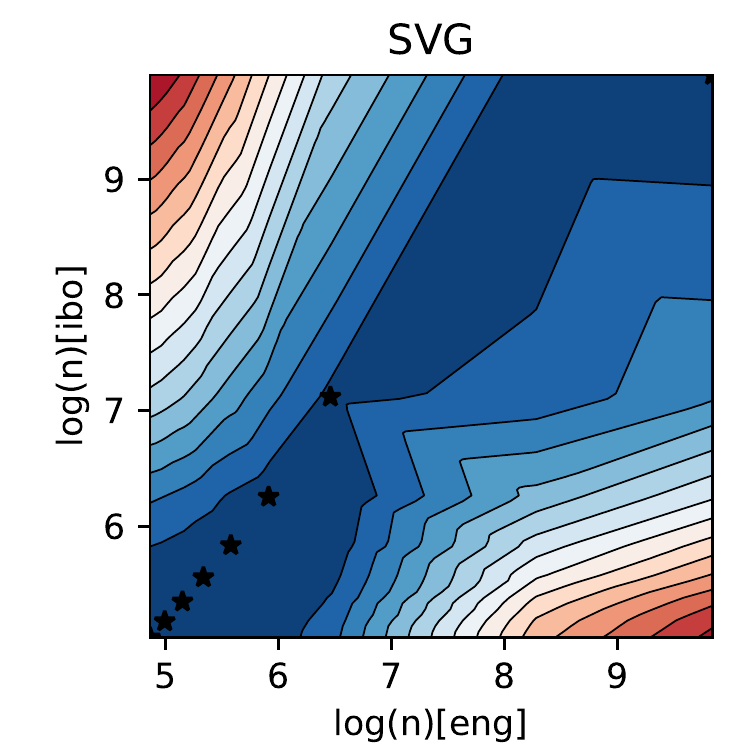}
		\includegraphics[width=\mywidth]{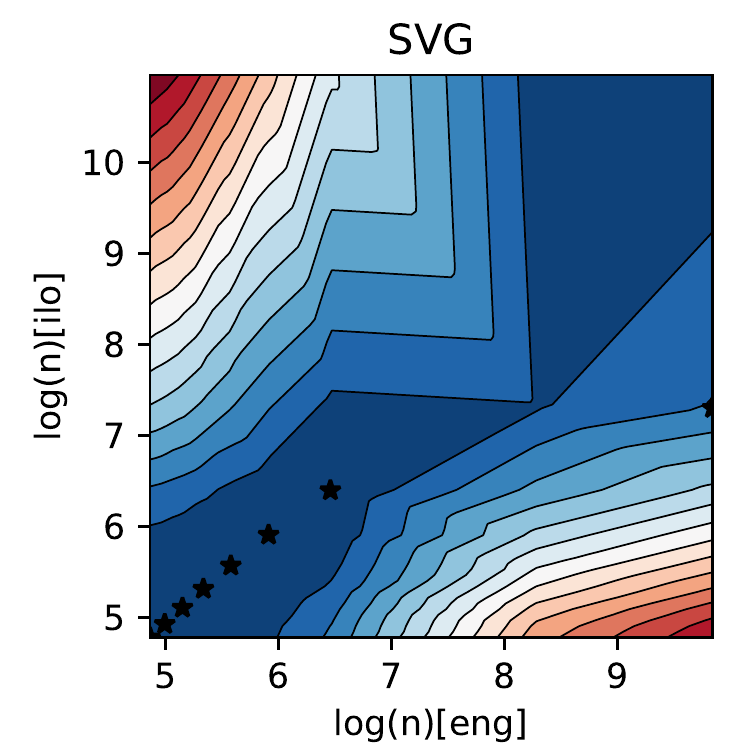}
		\includegraphics[width=\mywidth]{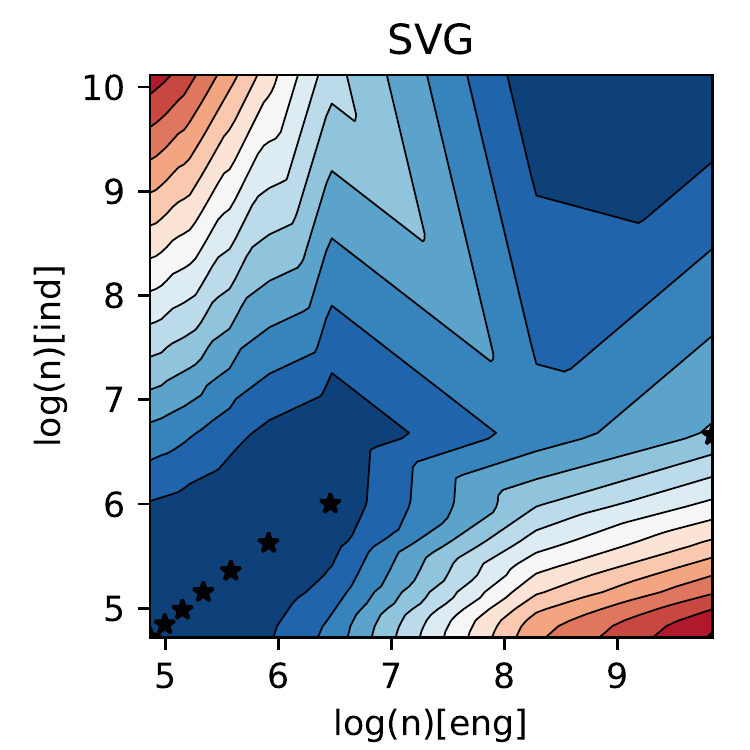}
		\includegraphics[width=\mywidth]{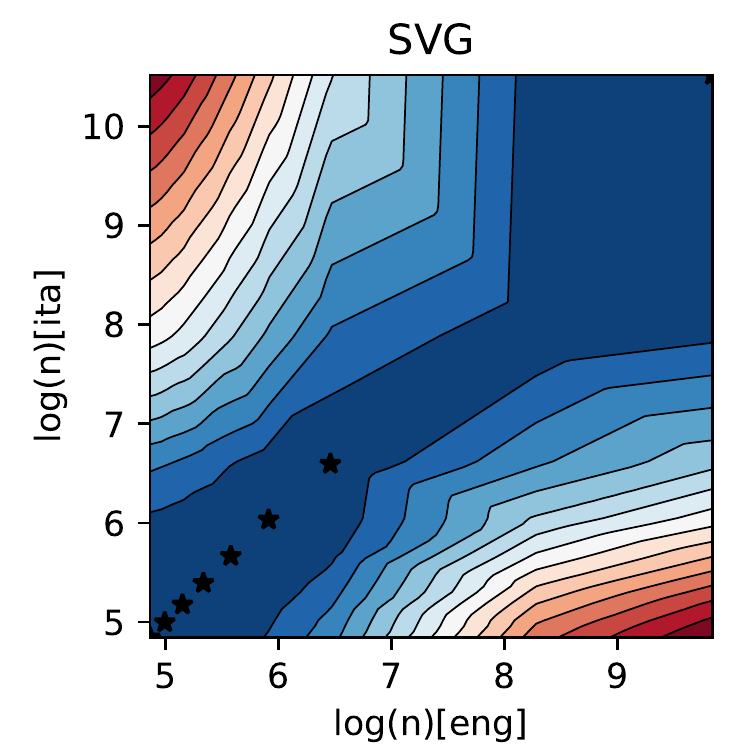}
		\includegraphics[width=\mywidth]{assets/plots3,12,svg/12,eng_newworld2013,jpn_newworld}
		\includegraphics[width=\mywidth]{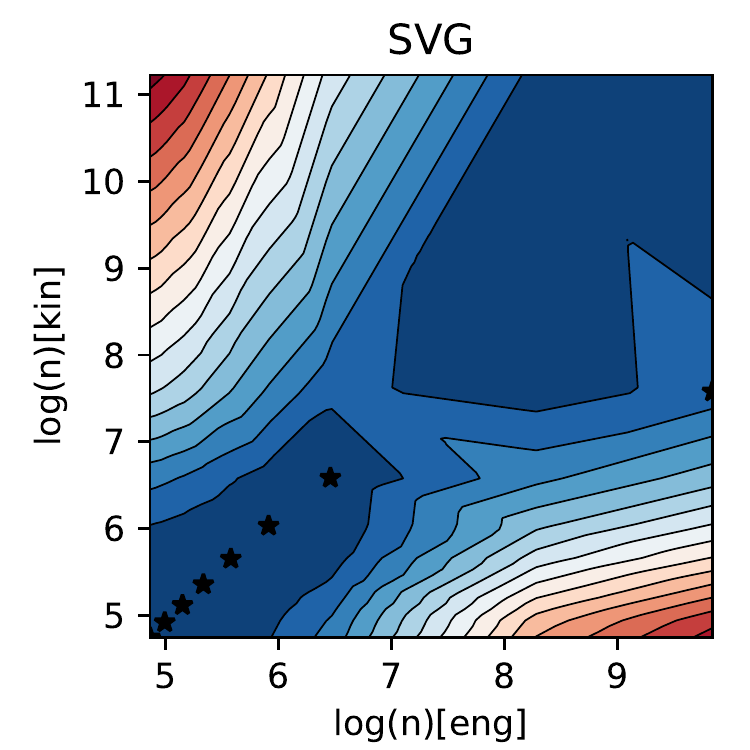}
		\includegraphics[width=\mywidth]{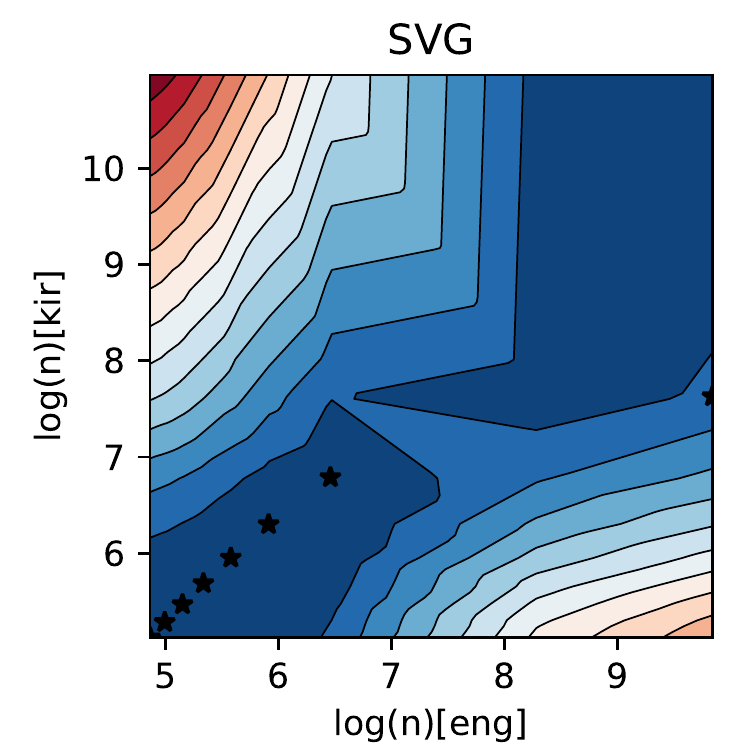}
		\includegraphics[width=\mywidth]{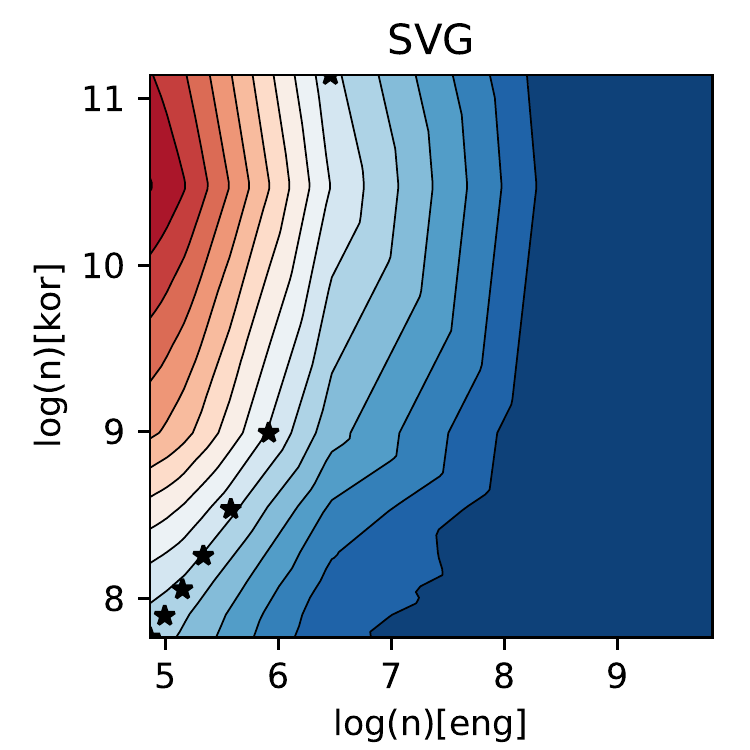}
		\includegraphics[width=\mywidth]{assets/plots3,12,svg/12,eng_newworld2013,kor_newworld2014}
		\includegraphics[width=\mywidth]{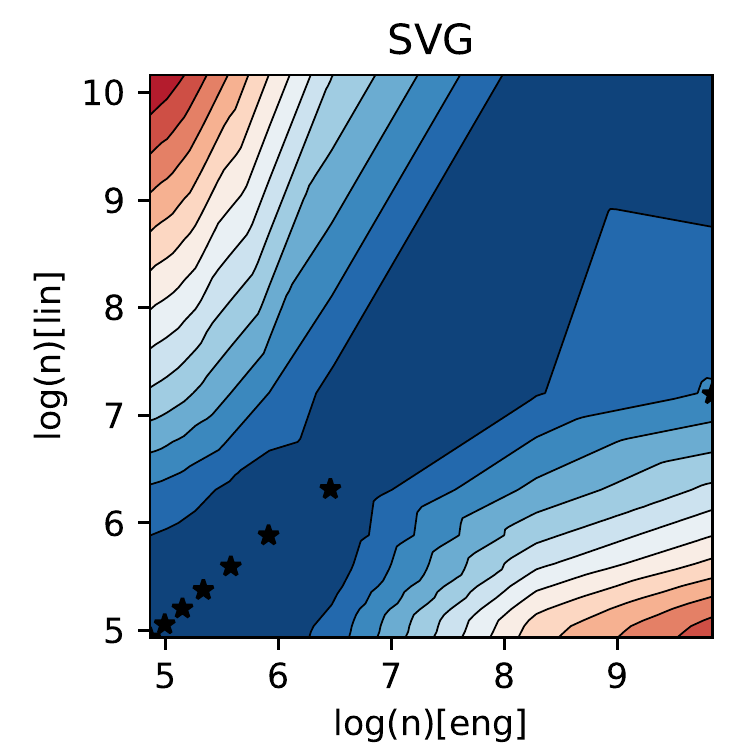}
		\includegraphics[width=\mywidth]{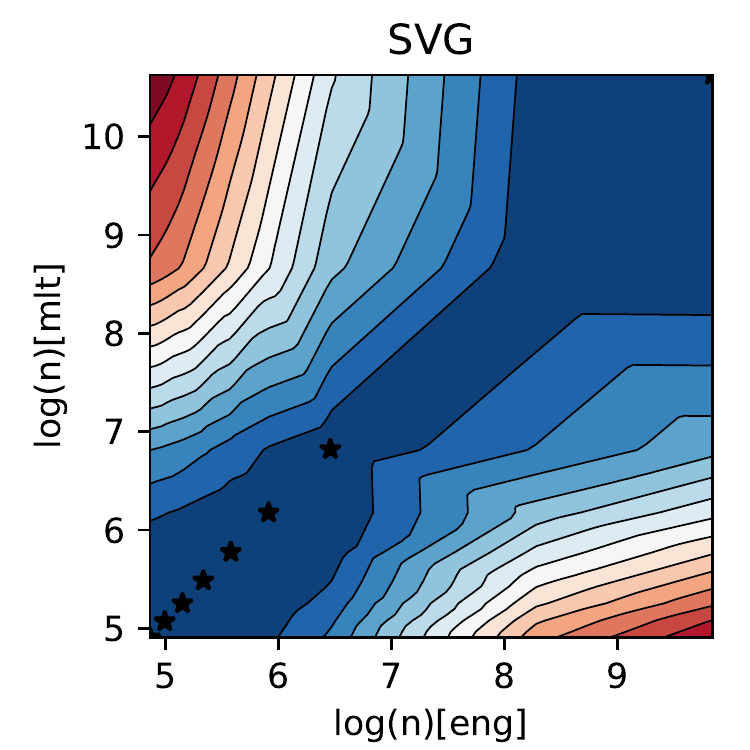}
		\includegraphics[width=\mywidth]{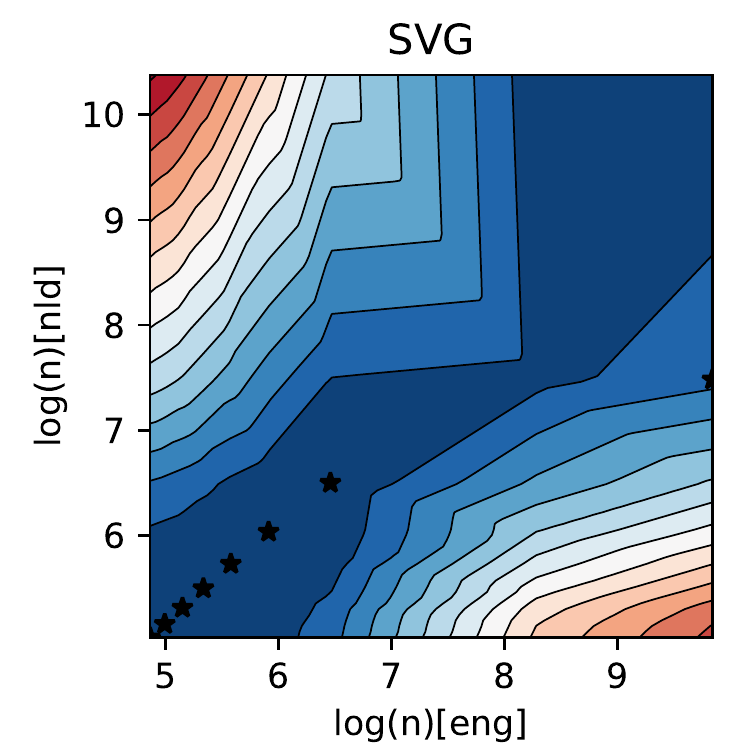}
		\includegraphics[width=\mywidth]{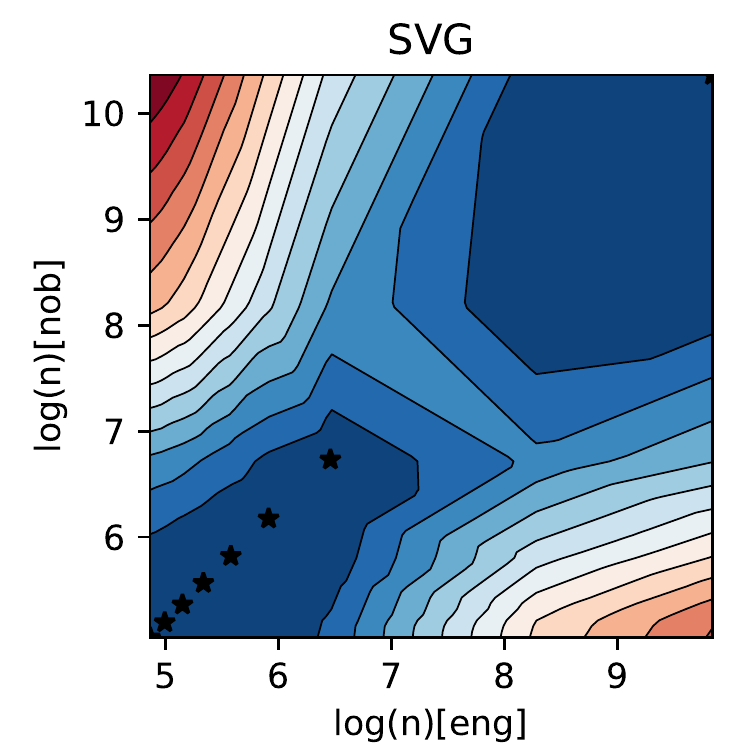}
		\includegraphics[width=\mywidth]{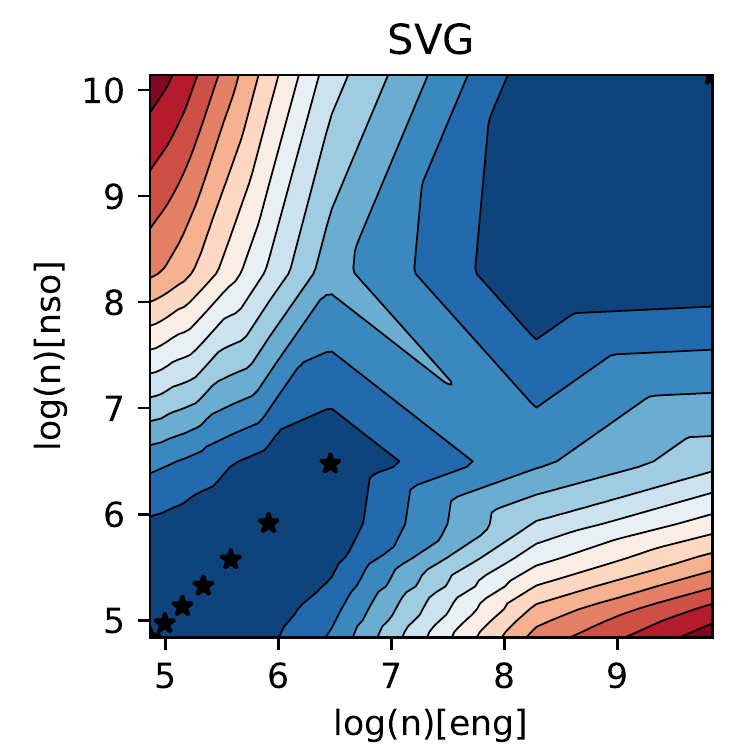}
		\includegraphics[width=\mywidth]{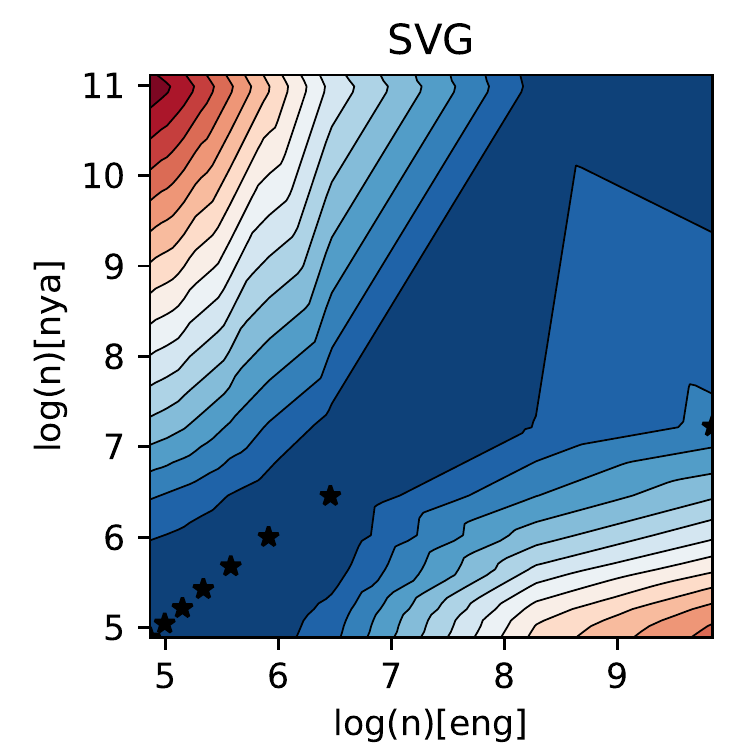}
		\includegraphics[width=\mywidth]{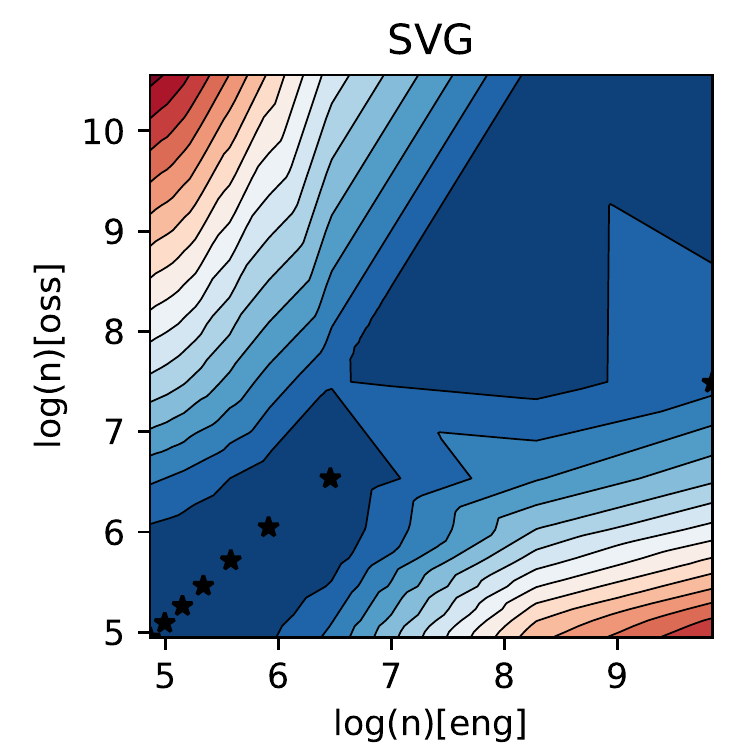}
		\includegraphics[width=\mywidth]{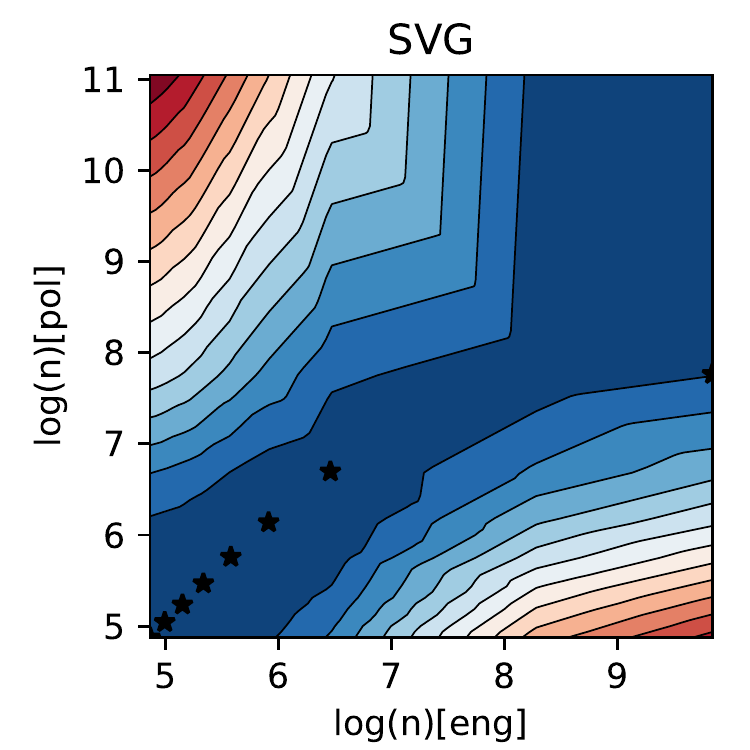}
		\includegraphics[width=\mywidth]{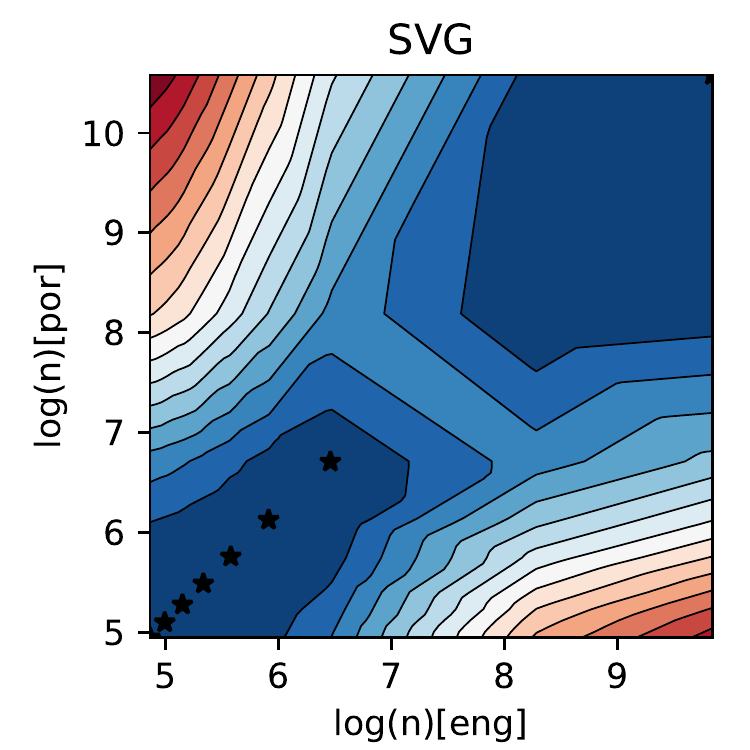}
		\includegraphics[width=\mywidth]{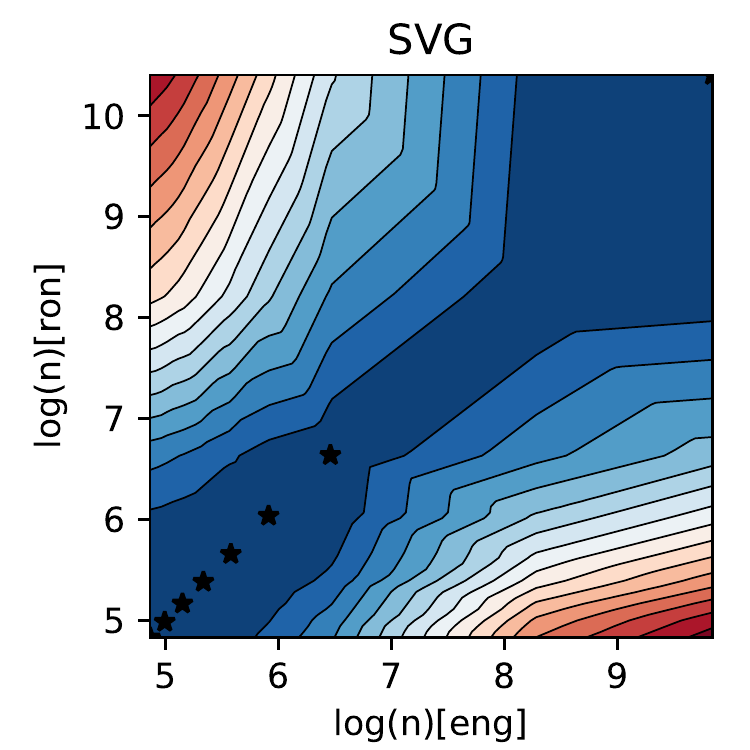}
		\caption{Figures showing the interpolated SVG for different language pairs as computed on the PBC (2/4).}
	\end{figure*}
	
	\begin{figure*}
		\centering
		\def\mywidth{0.24\textwidth}
		\centering
		\includegraphics[width=\mywidth]{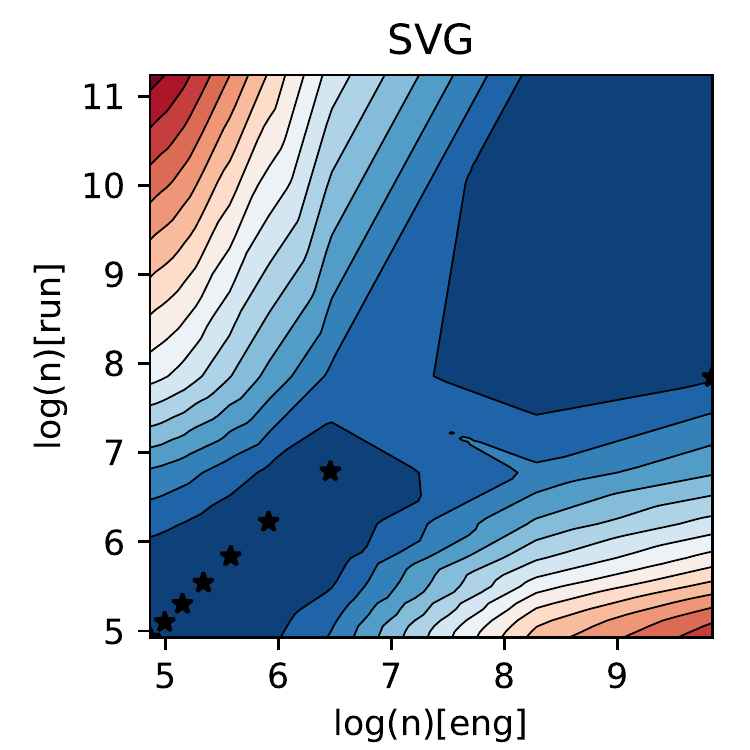}
		\includegraphics[width=\mywidth]{assets/plots3,12,svg/12,eng_newworld2013,rus_newworld}
		\includegraphics[width=\mywidth]{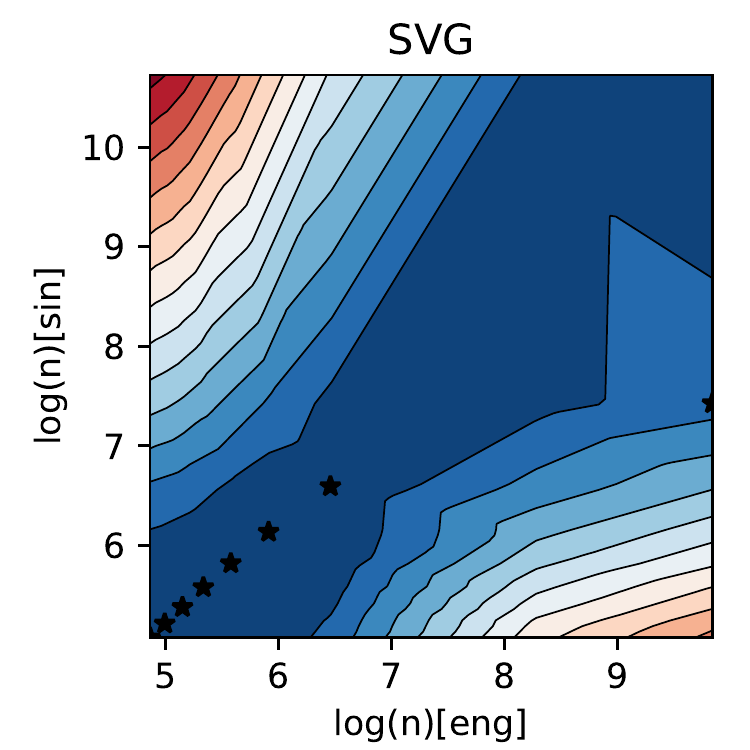}
		\includegraphics[width=\mywidth]{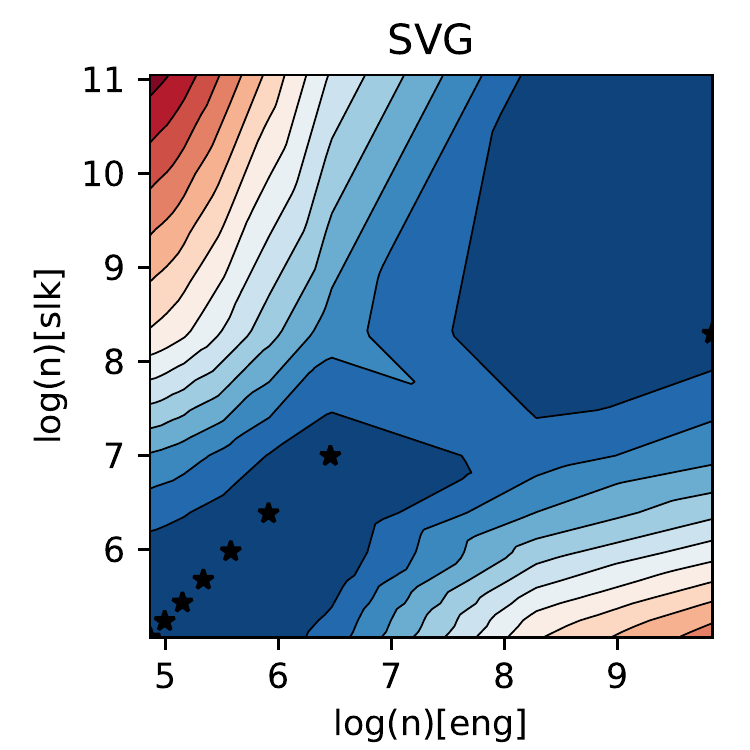}
		\includegraphics[width=\mywidth]{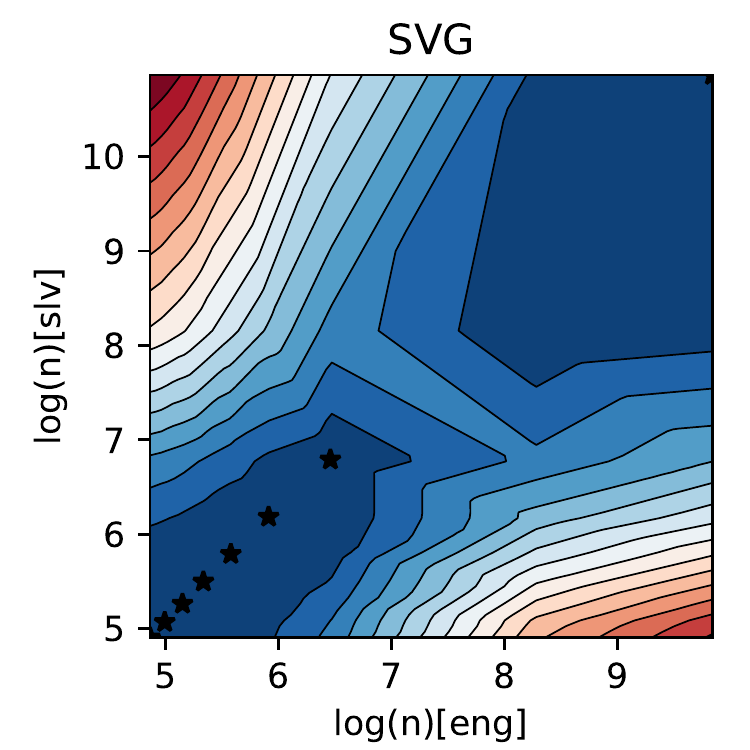}
		\includegraphics[width=\mywidth]{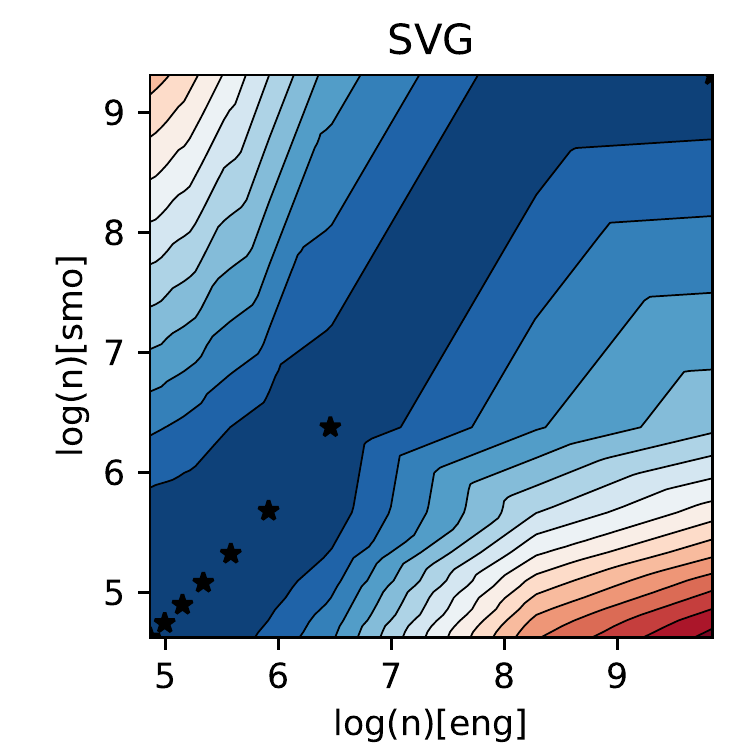}
		\includegraphics[width=\mywidth]{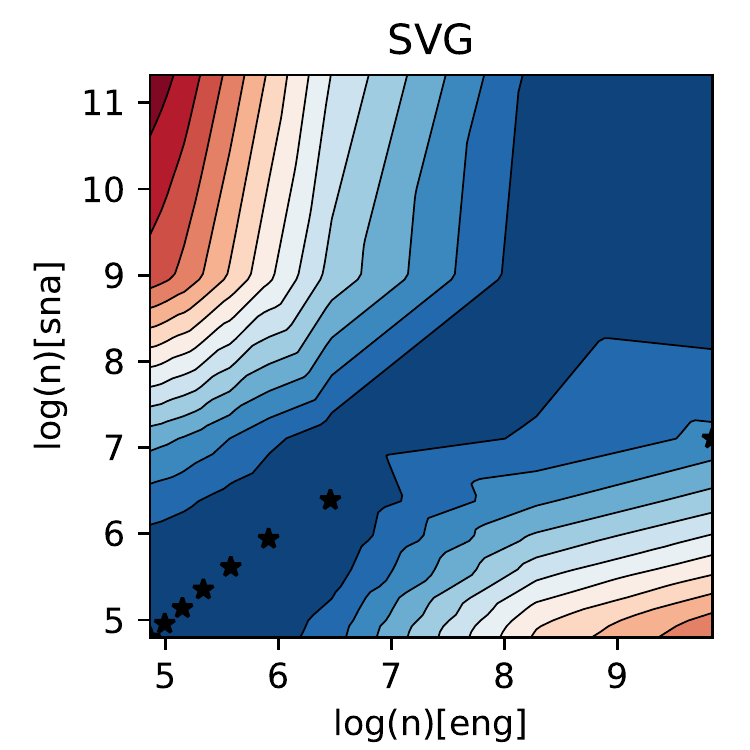}
		\includegraphics[width=\mywidth]{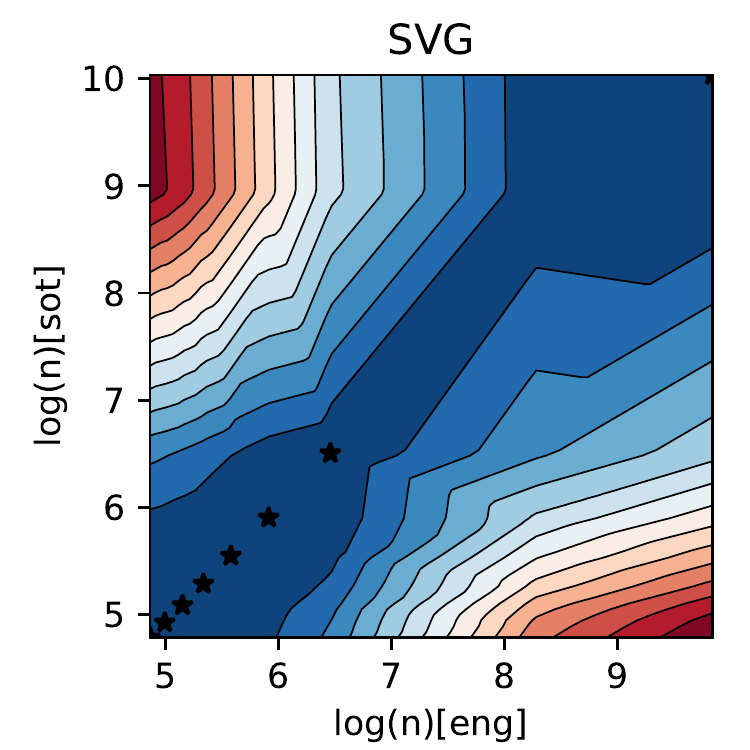}
		\includegraphics[width=\mywidth]{assets/plots3,12,svg/12,eng_newworld2013,spa_newworld}
		\includegraphics[width=\mywidth]{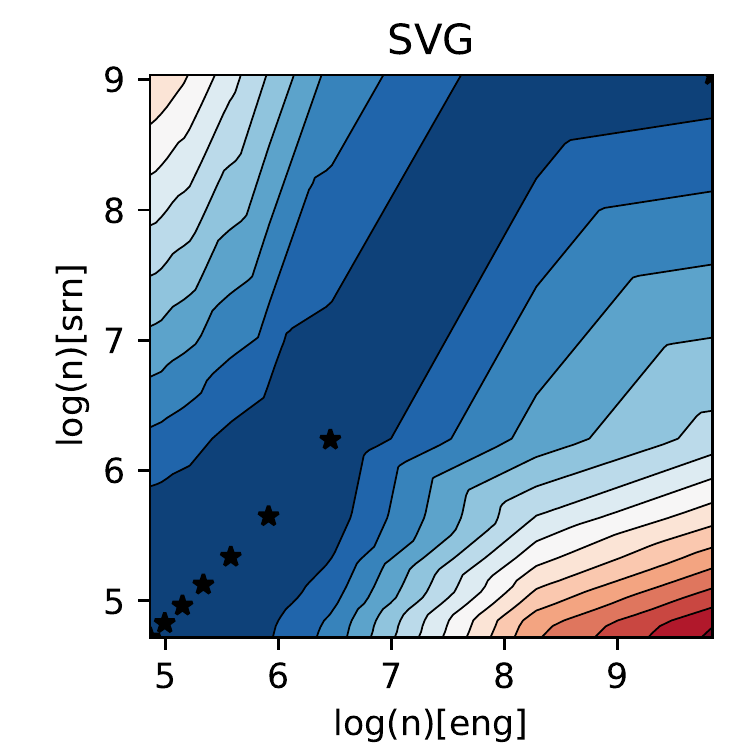}
		\includegraphics[width=\mywidth]{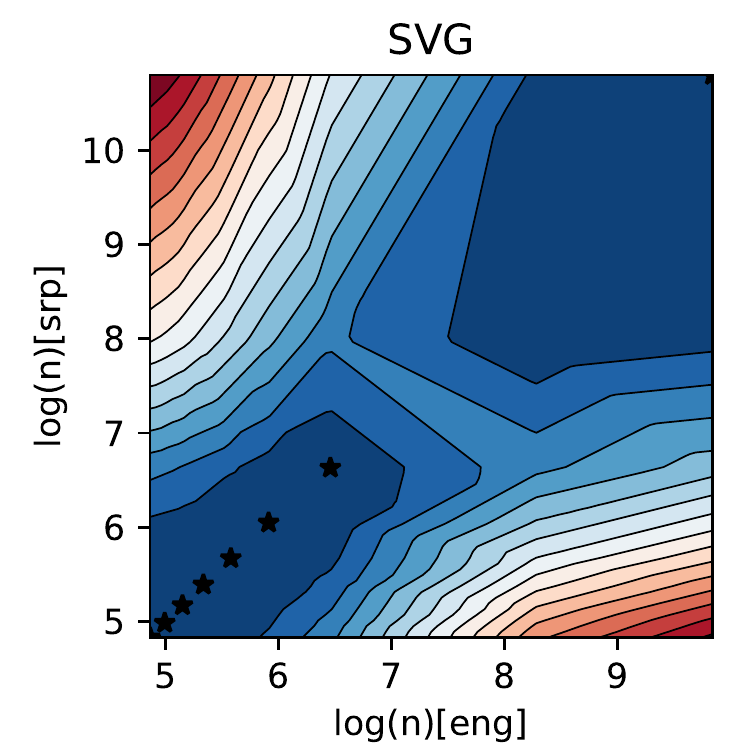}
		\includegraphics[width=\mywidth]{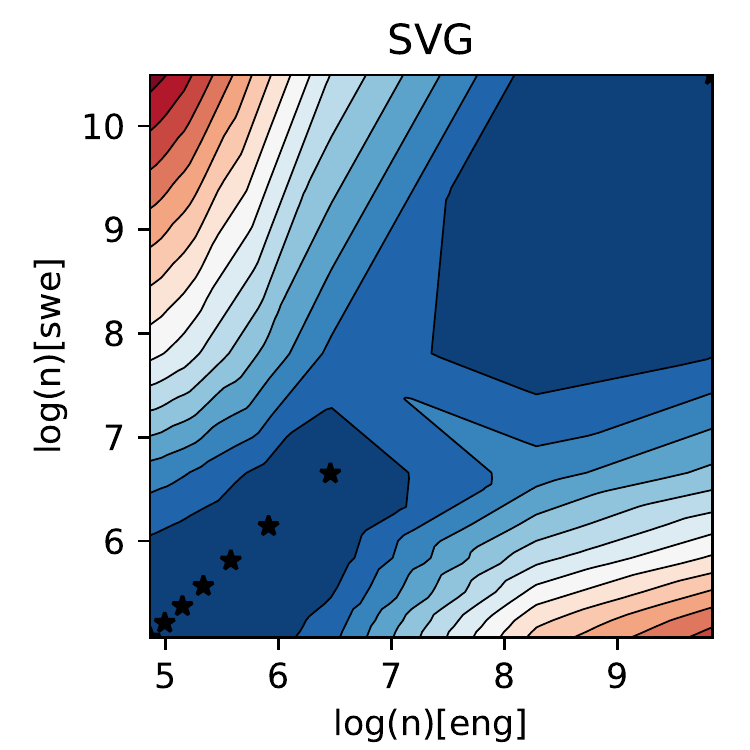}
		\includegraphics[width=\mywidth]{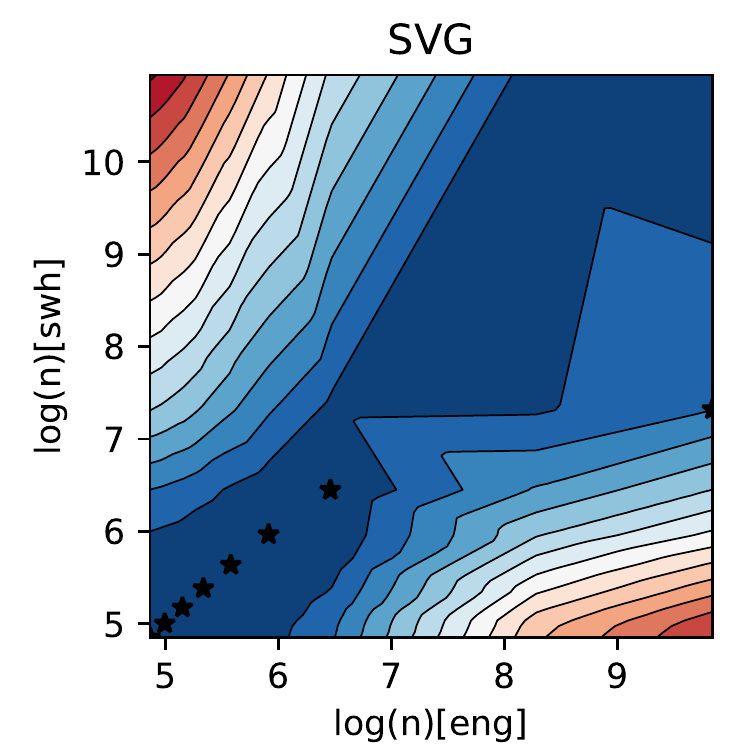}
		\includegraphics[width=\mywidth]{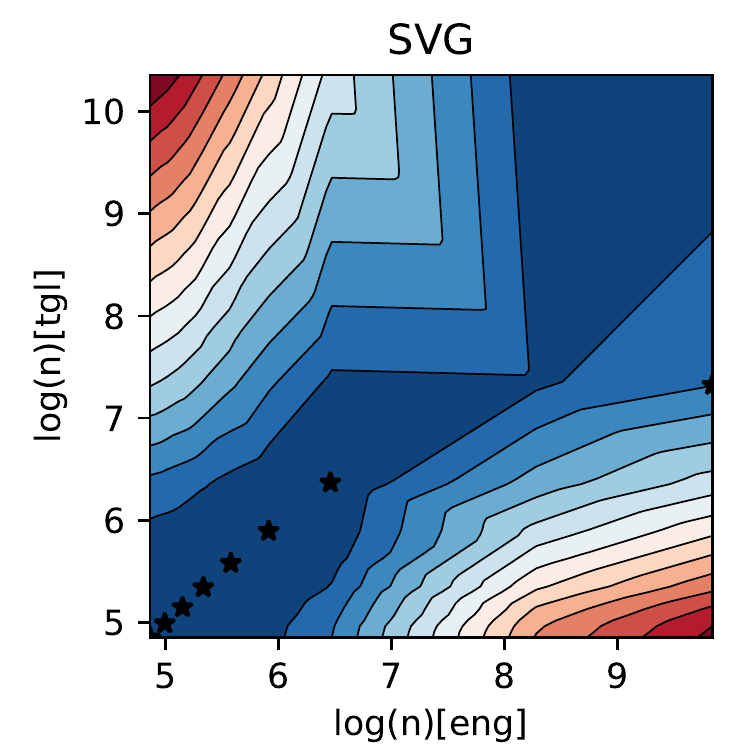}
		\includegraphics[width=\mywidth]{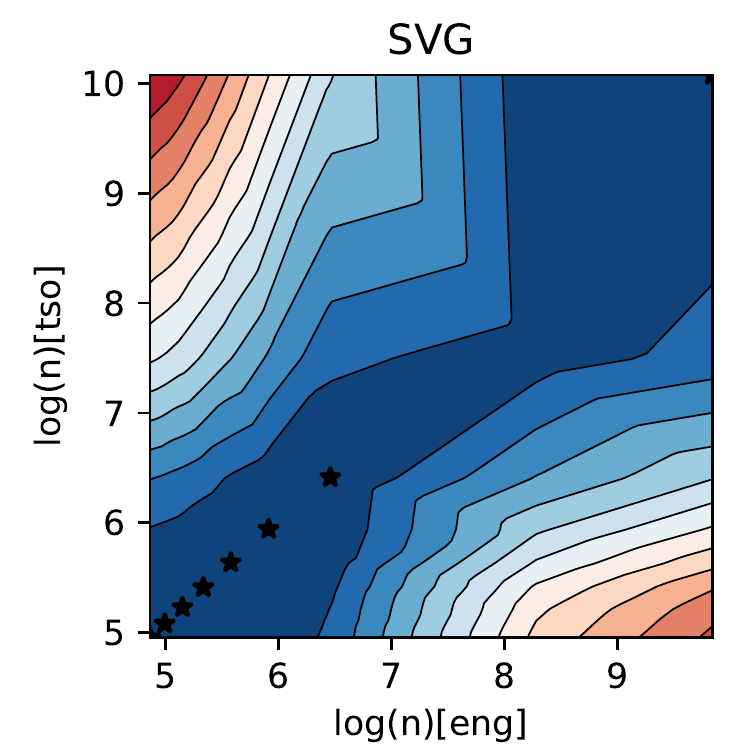}
		\includegraphics[width=\mywidth]{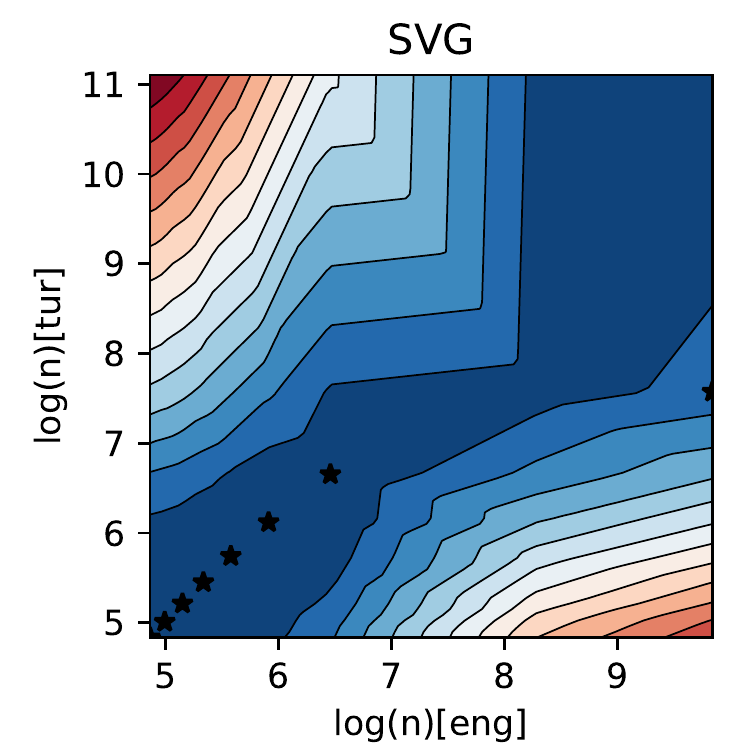}
		\caption{Figures showing the interpolated SVG for different language pairs as computed on the PBC (3/4).}
	\end{figure*}
	
	\begin{figure*}
		\centering
		\def\mywidth{0.24\textwidth}
		\centering
		\includegraphics[width=\mywidth]{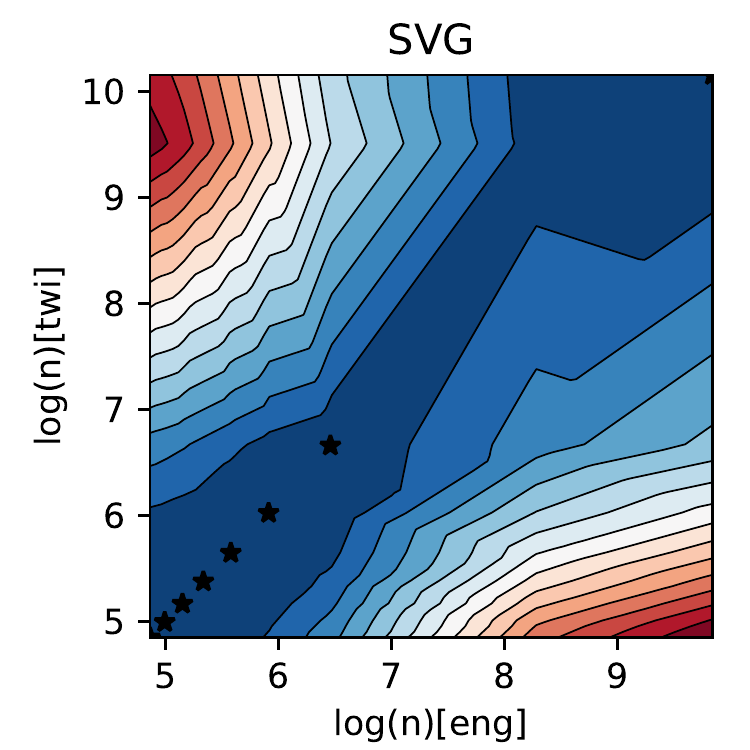}
		\includegraphics[width=\mywidth]{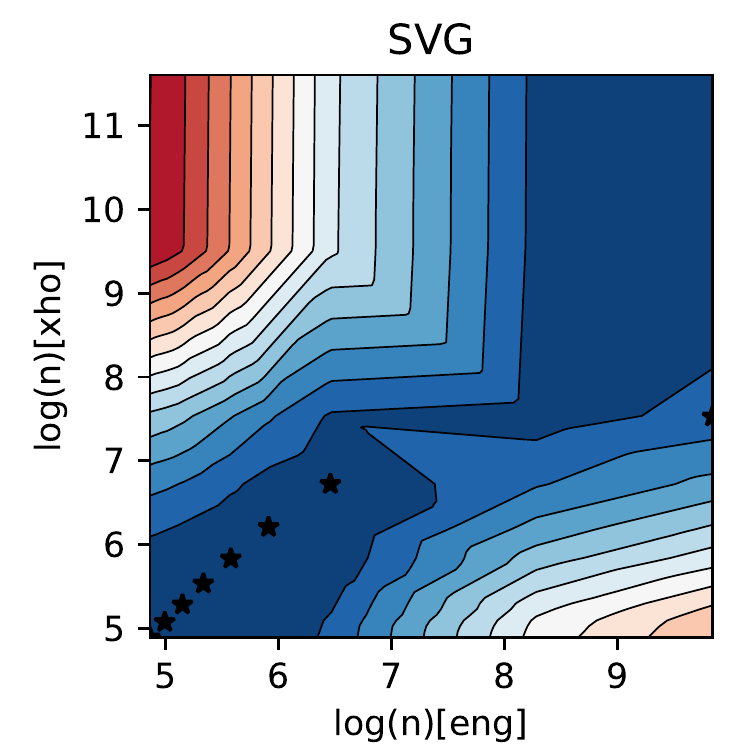}
		\includegraphics[width=\mywidth]{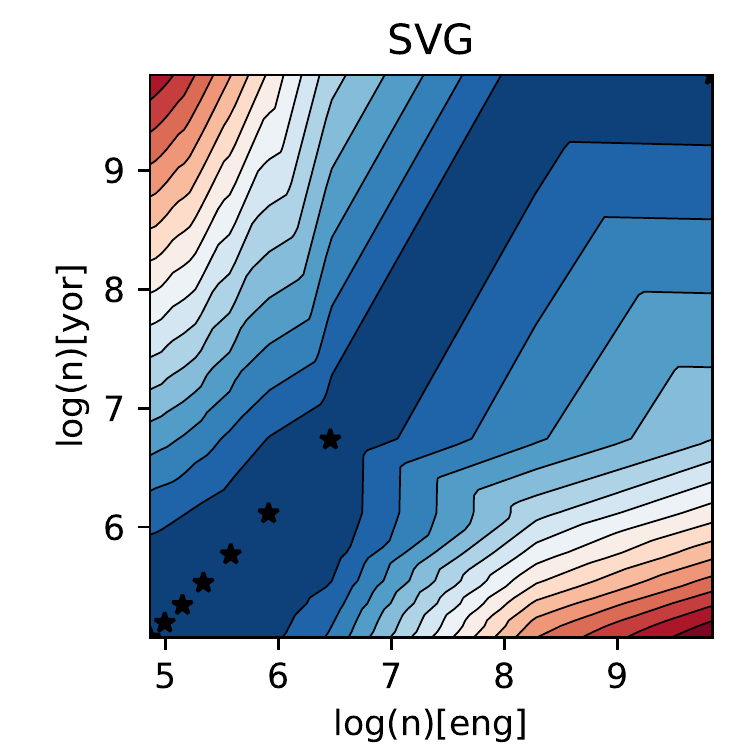}
		
		\includegraphics[width=\mywidth]{assets/plots3,12,svg/12,eng_newworld2013,zho_newworld}
		\includegraphics[width=\mywidth]{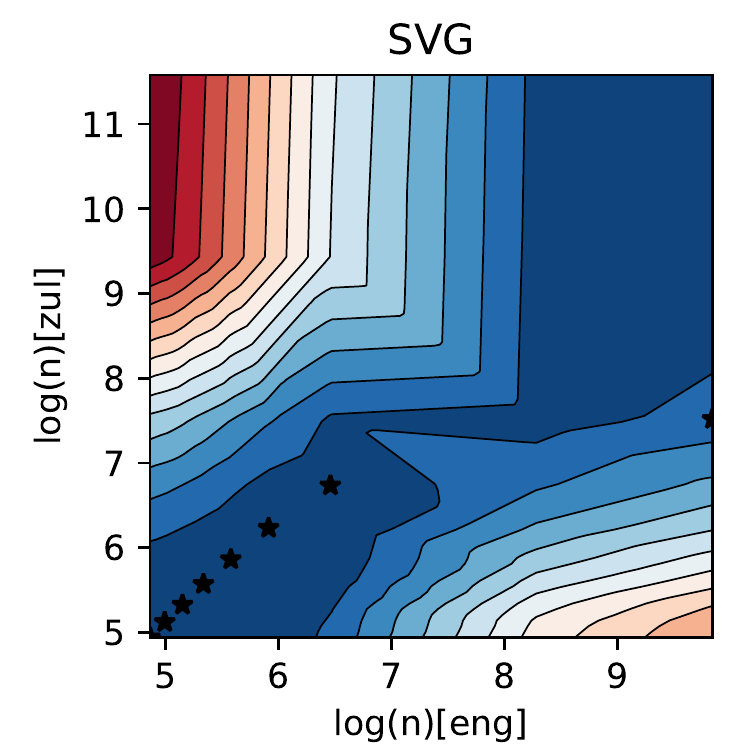}
		\caption{Figures showing the interpolated SVG for different language pairs as computed on the PBC (4/4).}
	\end{figure*}

\end{document}